\documentclass[10pt,twocolumn,letterpaper]{article}

\usepackage{3dv}
\usepackage{times}
\usepackage{epsfig}
\usepackage{graphicx}
\usepackage{amsmath}
\usepackage{amssymb}
\usepackage{booktabs}
\usepackage{breqn}
\usepackage{caption}
\usepackage{enumitem}
\usepackage{multirow}
\usepackage{subcaption}
\newcommand{\norm}[1]{\left\lVert#1\right\rVert}

\usepackage[pagebackref=true,breaklinks=true,letterpaper=true,colorlinks,bookmarks=false]{hyperref}

\usepackage{xcolor}

\usepackage{enumitem}
\setlist{nolistsep}
\threedvfinalcopy 

\def\threedvPaperID{102} 

\begin{document}

\title{PCN: Point Completion Network}

\author{Wentao Yuan \qquad Tejas Khot \qquad David Held \qquad Christoph Mertz \qquad Martial Hebert\\
The Robotics Institute\\
Carnegie Mellon University\\
{\tt\small \{wyuan1, tkhot, dheld, cmertz, hebert\}@cs.cmu.edu}}

\maketitle

\begin{abstract}
Shape completion, the problem of estimating the complete geometry of objects from partial observations, lies at the core of many vision and robotics applications. In this work, we propose Point Completion Network (PCN), a novel learning-based approach for shape completion. 
Unlike existing shape completion methods, PCN directly operates on raw point clouds without any structural assumption (e.g. symmetry) or annotation (e.g. semantic class) about the underlying shape. It features a decoder design that enables the generation of fine-grained completions while maintaining a small number of parameters.
Our experiments show that PCN produces dense, complete point clouds with realistic structures in the missing regions on inputs with various levels of incompleteness and noise, including cars from LiDAR scans in the KITTI dataset.
Code, data and trained models are available at \href{https://wentaoyuan.github.io/pcn}{https://wentaoyuan.github.io/pcn}.
\end{abstract}

\section{Introduction}

\begin{figure}
    \centering
    \includegraphics[width=\linewidth]{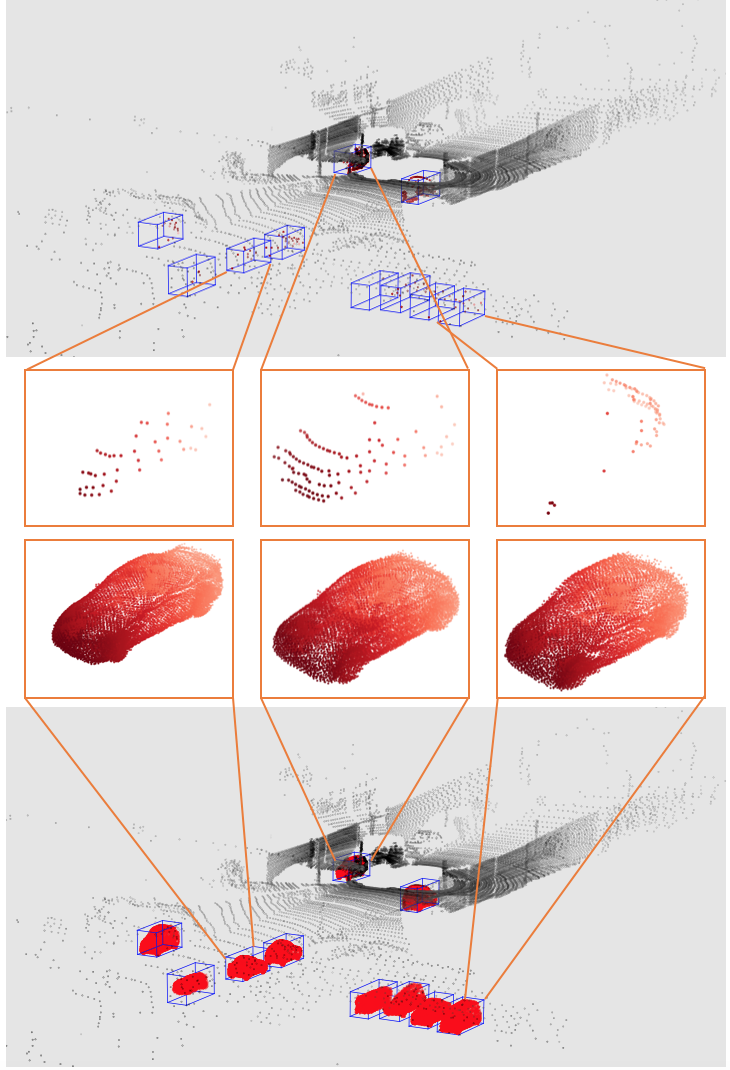}
    \caption{\textbf{(Top) Raw LiDAR scan from KITTI} \cite{geiger2013vision}. Note how the cars are barely recognizable due to incompleteness of the data. \textbf{(Bottom) Completed scan generated by PCN} on individual car point clouds segmented from the scene.}
    \label{fig:kitti}
\end{figure}

Real-world 3D data are often incomplete, causing loss in geometric and semantic information. For example, the cars in the LiDAR scan shown in Figure~\ref{fig:kitti} are hardly recognizable due to sparsity of the data points and missing regions caused by limited sensor resolution and occlusion. In this work, we present a novel learning-based method to complete these partial data using an encoder-decoder network that directly maps partial shapes to complete shapes, both represented as 3D point clouds.

Our work is inspired by a number of recent works \cite{dai2017shape,song2017semantic} which leverage large dataset of synthetic shapes to train deep neural networks that can infer the complete geometry from a single or a combination of partial views. However, a key difference between our approach and existing ones is in the representation for 3D data. The majority of existing methods voxelize the 3D data into occupancy grids or distance fields where convolutional networks can be applied. However, the cubically growing memory cost of 3D voxel grids limits the output resolution of these methods. Further, detailed geometry is often lost as an artifact of discretization. In contrast, our network is designed to operate on raw point clouds. This prevents the high memory cost and loss of geometric information caused by voxelization and allows our network to generate more fine-grained completions.

Designing a network that consumes and generates point clouds involves several challenges. First, a point cloud is an unordered set, which means permutations of the points do not change the geometry they represent. This necessitates the design of a feature extractor and a loss function that are permutation invariant. Second, there is no clear definition of local neighbourhoods in point clouds, making it difficult to apply any convolutional operation. Lastly, existing point cloud generation networks only generate a small set of points, which is not sufficient to capture enough detail in the output shape. Our proposed model tackles these challenges by combining a permutation invariant, non-convolutional feature extractor and a coarse-to-fine point set generator in a single network that is trained end-to-end.

The main contributions of this work are:
\begin{itemize}
    \item a learning-based shape completion method that operates directly on 3D point clouds without intermediate voxelization;
    \item a novel network architecture that generates a dense, complete point cloud in a coarse-to-fine fashion;
    \item extensive experiments showing improved completion results over strong baselines, robustness against noise and sparsity, generalization to real-world data and how shape completion can aid downstream tasks.
\end{itemize}

\section{Related Work}
\paragraph{3D Shape Completion}
Existing methods for 3D shape completion can be roughly categorized into geometry-based, alignment-based and learning-based approaches.

Geometry-based approaches complete shapes using geometric cues from the partial input without any external data. For example, surface reconstruction methods \cite{berger2014state, davis2002filling, nealen2006laplacian, sarkar2017learning, sorkine2004least, thanh2016field, zhao2007robust} generate smooth interpolations to fill holes in locally incomplete scans. Symmetry-driven methods \cite{mitra2006partial, mitra2013symmetry, pauly2008discovering, podolak2006planar, sipiran2014approximate, sung2015data, thrun2005shape} identify symmetry axes and repeating regular structures in the partial input in order to copy parts from observed regions to unobserved regions. These approaches assume moderately complete inputs where the geometry of the missing regions can be inferred directly from the observed regions. This assumption does not hold on most incomplete data from the real world.

Alignment-based approaches complete shapes by matching the partial input with template models from a large shape database. Some \cite{han2009bottom, li2015database, nan2012search, pauly2005example, shao2012interactive} retrieve the complete shape directly while some \cite{kalogerakis2012probabilistic, kim2013learning, martinovic2013bayesian, shen2012structure, sung2015data} retrieve object parts and then assemble them to obtain the complete shape. Other works \cite{blanz1999morphable, felzenszwalb2010object, gupta2015aligning, li2017shape, li2010analysis, rock2015completing} deform the retrieved model to synthesize shapes that are more consistent with the input. There are also works \cite{chauve2010robust, li2011globfit, nan2010smartboxes, schnabel2009completion,  yin2014morfit} that use geometric primitives such as planes and quadrics in place of a shape database. These methods require expensive optimization during inference, making them impractical for online applications. They are also sensitive to noise.

Learning-based approaches complete shapes with a parameterized model (often a deep neural network) that directly maps the partial input to a complete shape, which offers fast inference and better generalization. Our method falls into this category. While most existing learning-based methods \cite{dai2017shape, han2017high, sharma2016vconv, smith2017improved, stutzlearning, thanh2016field, varley2017shape, yang20183d} represents shapes using voxels, which are convenient for convolutional neural networks, our method uses point clouds, which preserve complete geometric information about the shapes while being memory efficient. One recent work \cite{litany2017deformable} also explores deformable meshes as the shape representation. However, their method assumes all the shapes are in correspondence with a common reference shape, which limits its applicability to certain shape categories such as humans or faces.






\paragraph{Deep Learning on Point Clouds}
Our method is built upon several recent advances in deep neural networks that operates on point clouds. PointNet and its extension \cite{qi2017pointnet, qi2017pointnet++} is the pioneer in this area and the state-of-the-art while this work was developed. It combines pointwise multi-layer perceptrons with a symmetric aggregation function that achieves invariance to permutation and robustness to perturbation, which are essential for effective feature learning on point clouds. Several alternatives \cite{li2018pointcnn, su2018splatnet, tatarchenko2018tangent, wang2018deep, wang2018dynamic} have been proposed since then. Any of these can be incorporated into our proposed model as the encoder.

There are relatively fewer works on decoder networks which generates point sets from learned features. \cite{achlioptas2017learning} uses a simple fully-connected decoder, while \cite{fan2017point} proposes a multi-branch decoder combining fully-connected and deconvolution layers. Recently, \cite{yang2017foldingnet} introduces an interesting decoder design which mimics the deformation of a 2D plane into a 3D surface. However, none of these methods generates more than 2048 points. Our model combines the advantages of these designs to generate higher resolution outputs in an efficient manner.

\section{Problem Statement}
Let $X$ be a set of 3D points lying on the observed surfaces of an object obtained by a single observation or a sequence of observations from a 3D sensor. Let $Y$ be a dense set of 3D points uniformly sampled from the observed and unobserved surfaces of the object. We define the shape completion problem as predicting $Y$ given $X$.
Note that under this formulation, $X$ is not necessarily a subset of $Y$ and there is no explicit correspondence between points in $X$ and points in $Y$, because they are independently sampled from the underlying object surfaces.


We tackle this problem using supervised learning. Leveraging a large-scale synthetic dataset where samples of $X$ and $Y$ can be easily acquired, we train a neural network to predict $Y$ directly from $X$. The network is generic across multiple object categories and does not assume anything about the structure of underlying objects such symmetry or planarity. The network architecture is described in Section~\ref{sec:arch} and the training process is described in Section~\ref{sec:training}.


\begin{figure*}[ht]
    \centering
    \includegraphics[width=\linewidth]{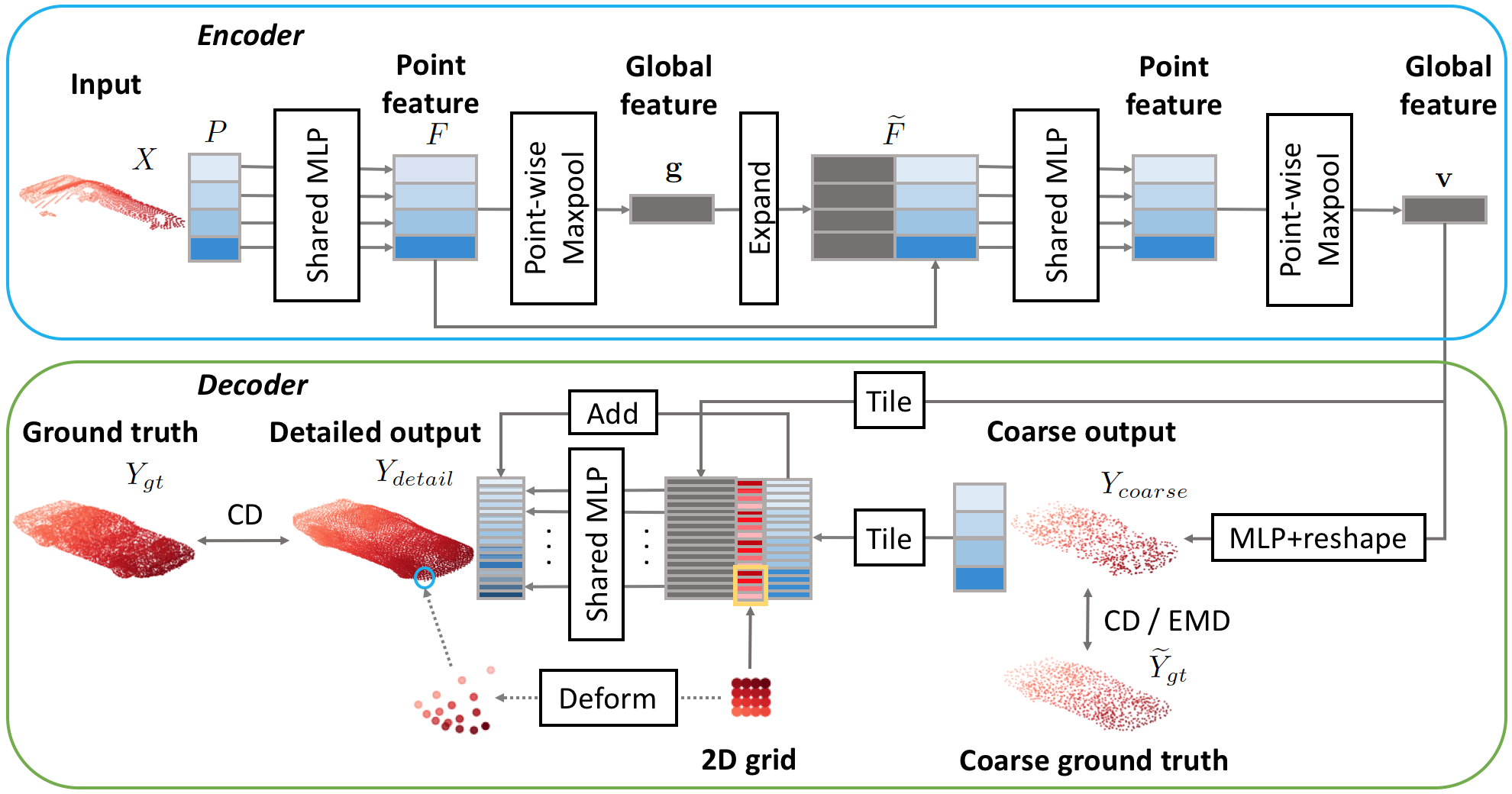}
    \caption{\textbf{PCN Architecture}. The encoder abstracts the input point cloud $X$ as a feature vector $\mathbf{v}$. The decoder uses $\textbf{v}$ to first generate a coarse output $Y_{coarse}$ followed by a detailed output $Y_{detail}$. Each colored rectangle denotes a matrix row. Same color indicates same content.}
    \label{fig:model}
\end{figure*}

\section{Point Completion Network} \label{sec:arch}
In this section, we describe the architecture of our proposed model, the Point Completion Network (PCN). As shown in Figure~\ref{fig:model}, PCN is an encoder-decoder network. The encoder takes the input point cloud $X$ and outputs a $k$-dimensional feature vector. The decoder takes this feature vector and produces a coarse output point cloud $Y_{coarse}$ and a detailed output point cloud $Y_{detail}$. The loss function $L$ is computed between the ground truth point cloud $Y_{gt}$ and the outputs of the decoder, which is used to train the entire network through backpropagation.
Note that, unlike an auto-encoder, we do not explicitly enforce the network to retain the input points in its output.
Instead, it learns a projection from the space of partial observations to the space of complete shapes. Next, we describe the specific design of the encoder, decoder and the loss function used.

\subsection{Point Feature Encoding}
The encoder is in charge of summarizing the geometric information in the input point cloud as a feature vector $\mathbf{v}\in\mathbb{R}^k$ where $k=1024$. Our proposed encoder is an extended version of PointNet \cite{qi2017pointnet}. It inherits the invariance to permutation and tolerance to noise from PointNet and can handle inputs with various number of points.

Specifically, the encoder consists of two stacked PointNet (PN) layers. The first layer consumes $m$ input points represented as an $m\times3$ matrix $P$ where each row is the 3D coordinate of a point $\mathbf{p}_i=(x,y,z)$. A shared multi-layer perceptron (MLP) consisting of two linear layers with ReLU activation is used to transform each $\mathbf{p}_i$ into a point feature vector $\mathbf{f}_i$. This gives us a feature matrix $F$ whose rows are the learned point features $\mathbf{f}_i$. Then, a point-wise maxpooling is performed on $F$ to obtain a $k$-dimensional global feature $\mathbf{g}$, where $\mathbf{g}_j=\max_{i=1,\dots,m}\{F_{ij}\}$ for $j=1,\dots,k$. The second PN layer takes $F$ and $\mathbf{g}$ as input. It first concatenates $\mathbf{g}$ to each $\mathbf{f}_i$ to obtain an augmented point feature matrix $\widetilde{F}$ whose rows are the concatenated feature vectors $[\mathbf{f}_i~\mathbf{g}]$. Then, $\widetilde{F}$ is passed through another shared MLP and point-wise max pooling similar to the ones in the first layer, which gives the final feature vector $\mathbf{v}$.


\subsection{Multistage Point Generation} \label{sec:decoder}
The decoder is responsible for generating the output point cloud from the feature vector $\mathbf{v}$. Our proposed decoder combines the advantages of the fully-connected decoder \cite{achlioptas2017learning} and the folding-based decoder \cite{yang2017foldingnet} in a multistage point generation pipeline. In our experiments, we show that our decoder outperforms either the fully-connected or the folding-based decoder on its own.

Our key observation is that the fully-connected decoder is good at predicting a sparse set of points which represents the global geometry of a shape. Meanwhile, the folding-based decoder is good at approximating a smooth surface which represents the local geometry of a shape. Thus, we divide the generation of the output point cloud into two stages. In the first stage, a coarse output $Y_{coarse}$ of $s$ points is generated by passing $\mathbf{v}$ through a fully-connected network with $3s$ output units and reshaping the output into a $s\times3$ matrix. In the second stage, for each point $\mathbf{q}_i$ in $Y_{coarse}$, a patch of $t=u^2$ points is generated in the local coordinates centered at $\mathbf{q}_i$ via the folding operation (refer to Section \ref{supp:folding} in the supplementary for details), and transformed into the global coordinates by adding $\mathbf{q}_i$ to the output. Combining all $s$ patches gives the detailed output $Y_{detail}$ consisting of $n=st$ points. This multistage process allows our network to generate a dense output point cloud with fewer parameters than fully-connected decoder (see Table \ref{tab:params}) and more flexibility than folding-based decoder. 


\subsection{Loss Function} \label{sec:loss}
The loss function measures the difference between the output point cloud and the ground truth point cloud. Since both point clouds are unordered, the loss needs to be invariant to permutations of the points. Two candidates of permutation invariant functions are introduced by \cite{fan2017point} -- Chamfer Distance (CD) and Earth Mover's Distance (EMD).

\begin{dmath}
CD(S_1, S_2) = \frac{1}{|S_1|}\sum_{x\in S_1}\min_{y\in S_2}\norm{x-y}_2 + \frac{1}{|S_2|}\sum_{y\in S_2}\min_{x\in S_1}\norm{y-x}_2 \label{eq:cd}
\end{dmath}

CD (\ref{eq:cd}) calculates the average closest point distance between the output point cloud $S_1$ and the ground truth point cloud $S_2$. We use the symmetric version of CD where the first term forces output points to lie close to ground truth points and the second term ensures the ground truth point cloud is covered by the output point cloud. Note that $S_1$ and $S_2$ need not be the same size to calculate CD.

\begin{equation}
    EMD(S_1, S_2) = \min_{\phi:S_1\to S_2}\frac{1}{|S_1|}\sum_{x\in S_1}\norm{x-\phi(x)}_2 \label{eq:emd}
\end{equation}

EMD (\ref{eq:emd}) finds a bijection $\phi:S_1\to S_2$ which minimizes the average distance between corresponding points. In practice, finding the optimal $\phi$ is too expensive, so we use an iterative $(1+\epsilon)$ approximation scheme \cite{bertsekas1985distributed}. Unlike CD, EMD requires $S_1$ and $S_2$ to be the same size.

\begin{dmath}
    L(Y_{coarse}, Y_{detail}, Y_{gt}) = d_1(Y_{coarse}, \widetilde{Y}_{gt}) + \alpha~d_2(Y_{detail}, Y_{gt}) \label{eq:loss}
\end{dmath}

Our proposed loss function (\ref{eq:loss}) consists of two terms, $d_1$ and $d_2$, weighted by hyperparameter $\alpha$. The first term is the distance between the coarse output $Y_{coarse}$ and the subsampled ground truth $\widetilde{Y}_{gt}$ which has the same size as $Y_{coarse}$. The second term is the distance between the detailed output $Y_{detail}$ and the full ground truth $Y_{gt}$.

In our experiments, we use both CD and EMD for $d_1$ but only CD for $d_2$. This is because the $O(n^2)$ complexity of the EMD approximation scheme makes it too expensive to compute during training when $n$ is large, while CD can be computed with $O(n\log n)$ complexity using efficient data structure for nearest neighbour search such as KDTree.

\section{Experiments}
In this section, we first describe the creation of a large-scale, multi-category dataset to train our model. Next, we compare our method against existing methods and ablated versions of our method on synthetic shapes. Finally, we show completion results on real-world point clouds and demonstrate how they can help downstream tasks such as point cloud registration.


\subsection{Data Generation and Model Training} \label{sec:training}
To train our model, we use synthetic CAD models from ShapeNet to create a large-scale dataset containing pairs of partial and complete point clouds $(X, Y)$. Specifically, we take 30974 models from 8 categories: airplane, cabinet, car, chair, lamp, sofa, table, vessel. The complete point clouds are created by sampling 16384 points uniformly on the mesh surfaces and the partial point clouds are generated by back-projecting 2.5D depth images into 3D. We use back-projected depth images for partial inputs instead of subsets of the complete point cloud in order to bring the input distribution closer to real-world sensor data. For each model, 8 partial point clouds are generated from 8 randomly distributed viewpoints. Note that the partial point clouds can have different sizes. 

We choose to use a synthetic dataset to generate training data because it contains complete, detailed 3D models of objects that are not available in real-world datasets. Despite the fact that recent datasets such as ScanNet \cite{dai2017scannet} or S3DIS \cite{armeni2017joint} have very high quality 3D reconstructions, these reconstructions have missing regions due to the limitations of the scanner's view, and thus are not good enough to use as ground truth for our model.

We reserve 100 models for validation and 150 models for testing. The rest is used for training. All our models are trained using the Adam \cite{kingma2014adam} optimizer with an initial learning rate of 0.0001 for 50 epochs and a batch size of 32. The learning rate is decayed by 0.7 every 50K iterations.

\begin{figure*}[ht]
  \begin{subfigure}[b]{0.66\textwidth}
    \includegraphics[width=\textwidth]{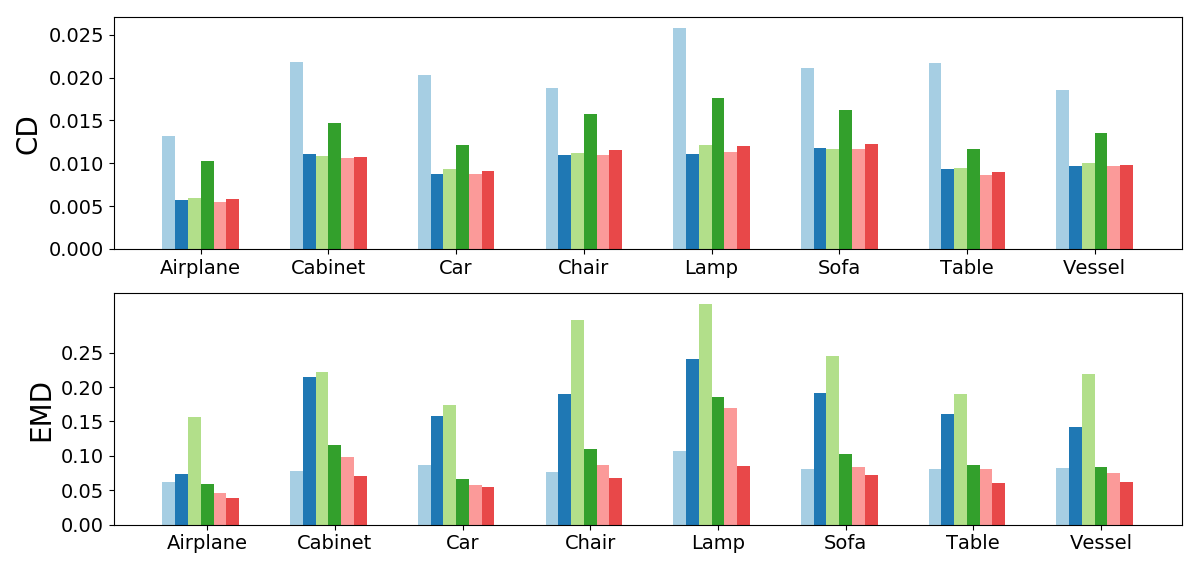}
    \caption{Results on trained categories}
    \label{fig:comparison1}
  \end{subfigure}
  \hfill
  \begin{subfigure}[b]{0.22\textwidth}
    \includegraphics[width=\textwidth]{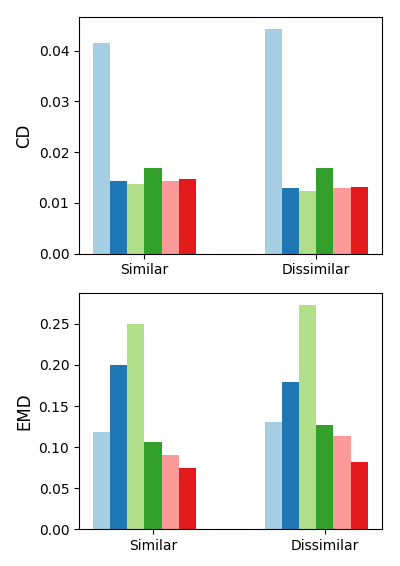}
    \caption{Results on novel categories}
    \label{fig:comparison2}
  \end{subfigure}
  \raisebox{1.2cm}{\includegraphics[width=0.1\textwidth]{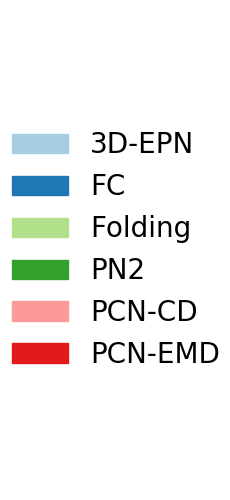}}
  \caption{\textbf{Quantitative comparison on ShapeNet}. For both CD (top) and EMD (below), lower is better. (a) shows results for test instances from the same categories used in training. (b) shows results for test instances from categories not included in training, which are divided into similar (bus, bed, bookshelf, bench) and dissimilar (guitar, motorbike, skateboard, pistol).}
  \label{fig:comparison}
\end{figure*}

\begin{figure*}[ht]
  \begin{subfigure}[b]{0.33\textwidth}
    \includegraphics[width=\textwidth]{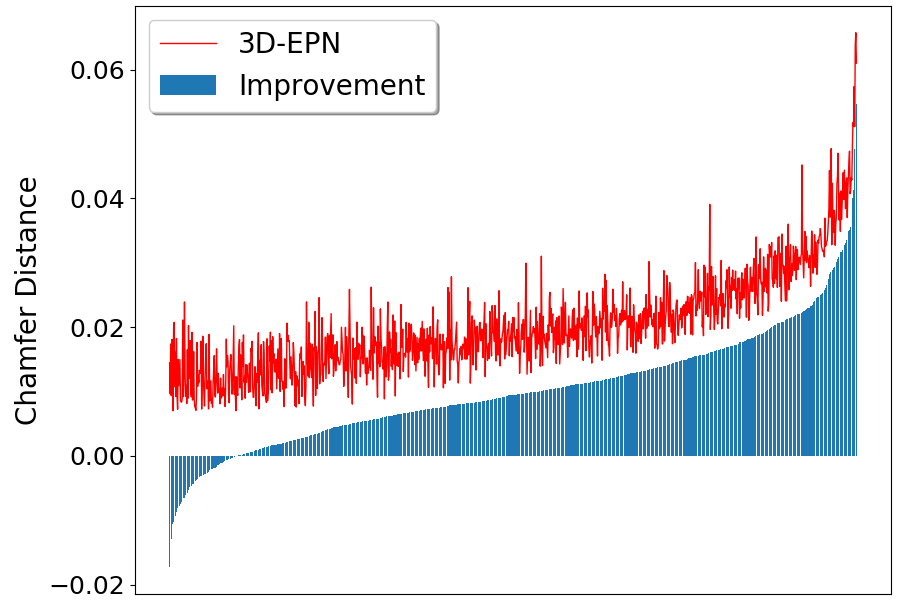}
    \caption{Chamfer Distance (CD)}
    \label{fig:cd_improve}
  \end{subfigure}
  \hfill
  \begin{subfigure}[b]{0.33\textwidth}
    \includegraphics[width=\textwidth]{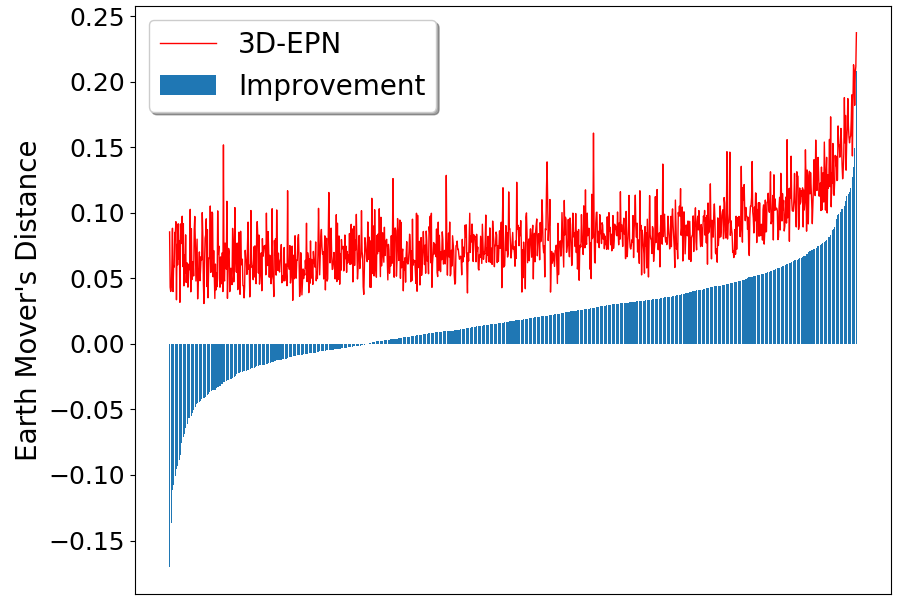}
    \caption{Earth Mover's Distance (EMD)}
    \label{fig:emd_improve}
  \end{subfigure}
  \hfill
  \begin{subfigure}[b]{0.33\textwidth}
    \includegraphics[width=\textwidth]{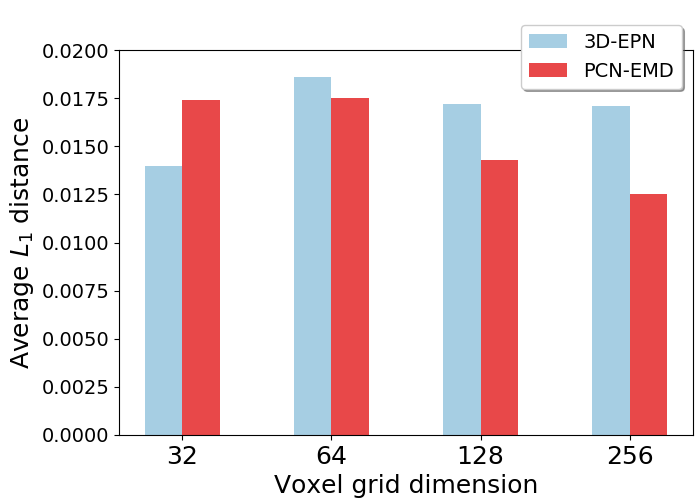}
    \caption{$L_1$ distance}
    \label{fig:voxel_comp}
  \end{subfigure}
  \caption{\textbf{Comparison between PCN-EMD and 3D-EPN}. (a) and (b) shows comparison of point cloud outputs. The $x$-axis represents different object instances. The height of the blue bar indicates the amount of improvement of PCN-EMD over 3D-EPN. The red curve is the error of 3D-EPN and the difference between the red curve and the blue bar is the error of PCN-EMD. PCN-EMD improves on the majority of instances. (c) shows comparison of distance field outputs. On the $y$-axis is the average $L_1$ distance on occluded voxels between output and ground truth distance fields. PCN-EMD achieves lower $L_1$ distance on higher resolutions.}
  \label{fig:cd_emd}
\end{figure*}

\subsection{Completion Results on ShapeNet} \label{sec:shapenet}
In this subsection, we compare our method against several strong baselines, including a representative volumetric network and modified versions of our model, on synthetic point clouds from ShapeNet. We also test the generalizability of these methods to novel shapes and the robustness of our model against occlusion and noise.

\paragraph{Baselines}
Previous point-based completion methods either assume more complete inputs than we have \cite{kazhdan2013screened} or prior knowledge of the shape such as semantic class, symmetry and part segmentation \cite{sung2015data}, and thus are not directly comparable to our method. Here, we compare our model against four strong baselines which, like our method, work on objects from multiple categories with different levels of incompleteness.

1) \textbf{3D-EPN} \cite{dai2017shape}: This is a representative of the class of volumetric completion methods that is also trained end-to-end on large synthetic dataset. To compare the distance field outputs of 3D-EPN with the point cloud outputs of PCN, we convert the distance fields into point clouds by extracting the isosurface at a small value $d$ and uniformly sampling 16384 points on the resulting mesh. To ensure fair comparison, we also convert the point cloud outputs of PCN into distance fields by calculating the distance from grid centers to the closest point in the output. 

2) \textbf{FC}: This is a network that uses the same encoder as PCN but the decoder is a 3-layer fully-connected network which directly outputs the coordinates of 16384 points.

3) \textbf{Folding}: This is a network that also uses the same encoder as PCN but the decoder is purely folding-based \cite{yang2017foldingnet}, which deforms a 128-by-128 2D grid into a 3D point cloud.

4) \textbf{PN2}: This is a network that uses the same decoder as our proposed model but the encoder is PointNet++ \cite{qi2017pointnet++}.

We provide two versions of our model for comparison, PCN-CD and PCN-EMD. The number of points in the coarse and detailed outputs are $s=1024$ and $n=16384$ respectively. For the loss on coarse output, PCN-CD uses CD and PCN-EMD uses EMD. Note that both models use CD for the loss on detailed output due to the computational complexity of EMD.

\paragraph{Test Set}
We created two test sets: one consists of 150 reserved shapes from the 8 object categories on which the models are trained; the other consists of 150 models from 8 novel categories that are not in the training set. We divide the novel categories into two groups: one that is visually similar to the training categories -- bed, bench, bookshelf and bus, and another that is visually dissimilar to the training categories -- guitar, motorbike, pistol and skateboard. The quantitative comparisons are shown in Figure~\ref{fig:comparison} and some qualitative examples are shown in Figure~\ref{fig:qualitative}.

\begin{figure}[ht]
    \centering
    \includegraphics[width=0.9\linewidth]{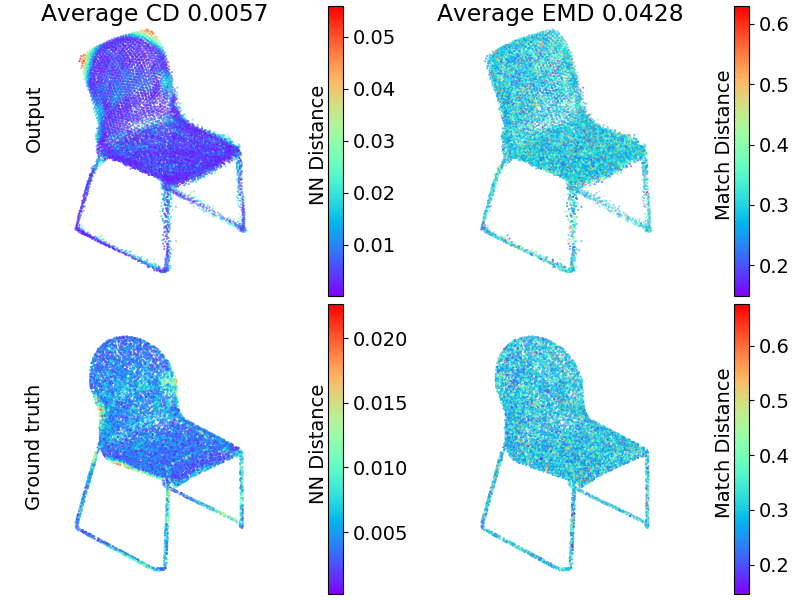}
    \caption{\textbf{Illustration of CD (left) and EMD (right)}. The top row shows the output of our model and the bottom row shows the ground truth. The points on the left are colored by their distance to the closest point in the other point cloud (nearest neighbor (NN) distance). The points on the right are colored by their distance to the corresponding point under the optimal bijection (match distance). Average CD is the mean of the NN distances and average EMD is the mean of the match distances.}
    \label{fig:metric}
\end{figure}

\paragraph{Metrics}
The metrics we use on point clouds are CD and EMD between the output and ground truth point clouds, as defined in \ref{sec:loss}. An illustration of the difference between the two metrics is shown in Figure~\ref{fig:metric}. We can see that CD is high where the global structure is different, e.g. around the corners of the chair back. On the other hand, EMD is more evenly distributed, as it penalizes density difference between the two point clouds. Note that on average, EMD is much higher than CD. This is because EMD requires one-to-one correspondences between the points, whereas the point correspondences used by CD can be one-to-many.

The metric we use on distance fields is the $L_1$ distance between the output and ground truth distance fields, same as in \cite{dai2017scannet}. To have comparable numbers across different dimensions, we convert the error from the voxel distance to distance in the model's metric space.

\paragraph{Generalizability to Novel Objects}
As shown in Figure~\ref{fig:comparison2}, our method outperforms all baselines on object from novel categories. More importantly, our model's performance is not significantly affected even on visually dissimilar categories (e.g. the pistol in Figure~\ref{fig:qualitative}). 
This shows the generality of the shape prior learned by our model.

\paragraph{Comparison to Volumetric Method}
It can be seen that our method outperforms 3D-EPN by a large margin on both CD and EMD. To better interpret the numbers, in Figures~\ref{fig:cd_improve},~\ref{fig:emd_improve}, we show the amount of improvement of our completion results over that of 3D-EPN on CD and EMD for each instance in the test set. The results of our method improve on the majority of instances. Further, they improve the most on examples where the error of 3D-EPN is high, indicating its ability to handle challenging cases where previous methods fail.

In Figure~\ref{fig:voxel_comp}, we show that our method achieves lower $L_1$ distance when its outputs are converted to a distance field. Moreover, the improvement of our method over 3D-EPN is more significant at higher resolutions.

\paragraph{Decoder Comparison}
The results in Figure~\ref{fig:comparison} show how our proposed decoder compares with existing decoder designs. Our multistage design which combines the advantages of fully-connected and folding-based decoders outperforms either design on its own. From the qualitative results, we observe that the fully-connected decoder does not have any constraints on the local density of the output points, and thus the output points are often over-concentrated in areas such as table top, which results in high EMD. On the other hand, the folding-based decoder often produces points that are floating in space and not consistent with the global geometry of the ground truth shape, which results in high CD. This is because the shapes in our dataset contain many concavities and sharp edges, which makes globally folding a 2D plane into a 3D shape very challenging. FoldingNet \cite{yang2017foldingnet} addresses this by chaining two folding operations. However, by only doing the folding operation locally, our decoder is able to achieve better performance with only one folding operation.

\paragraph{Encoder Comparison}
Another pair of comparison shown in Figure~\ref{fig:comparison} is between the stacked PN and PN2 \cite{qi2017pointnet++} as the encoder. PN2 is a representative of the class of hierarchical feature extraction networks that aggregate local information before global pooling. Our results show that it is outperformed by our stacked PN encoder which uses only global pooling. We believe this is because local pooling is less stable than global pooling due to suboptimality in the selection of local neighbourhoods for the partial data we are dealing with. Thus, we argue that stacking PN layers instead of doing local pooling is a better way of mixing local and global information.


\paragraph{Number of Parameters}
As shown in Table \ref{tab:params}, our model has an order of magnitude fewer parameters than 3D-EPN and FC while achieving significantly better performance.

\begin{table}[ht]
    \centering
    \caption{Number of trainable model parameters}
    \begin{small}
    \begin{tabular}{l|c|c|c|c|c}
        \toprule
        Method & 3D-EPN & FC & Folding & PN2 & Ours \\
        \midrule
        \# Params & 52.4M & 53.2M & 2.40M & 6.79M & 6.85M \\
        \bottomrule
    \end{tabular}
    \end{small}
    \label{tab:params}
\end{table}

\paragraph{Robustness to occlusion and noise}
Now, we test the robustness of our method to sensor noise and large occlusions. Specifically, we perturbed the depth map with Gaussian noise whose standard deviation is 0.01 times the scale of the depth measurements, and occluded it with a mask that covers $p$ percent of points, where $p$ ranges from $0\%$ to $80\%$. Additionally, we randomly set $1\%$ of the measurements to $d_{max}=1.6$.

\begin{figure}[ht]
    \centering
    \begin{flushright}
    \includegraphics[width=\linewidth]{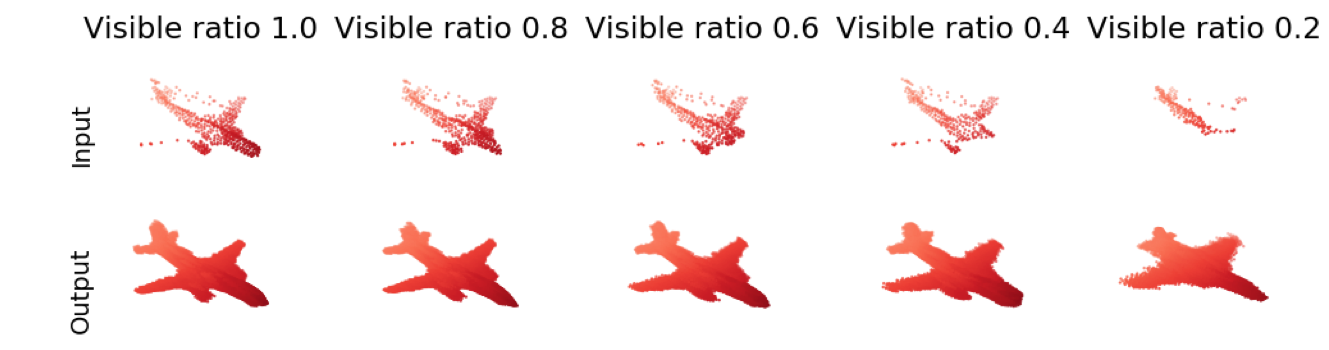}
    \end{flushright}
    \includegraphics[width=\linewidth]{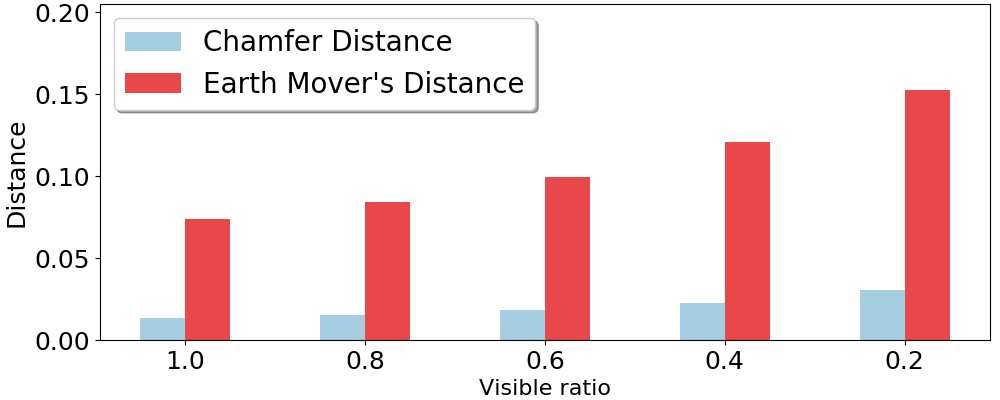}
    \caption{Qualitative (top) and quantitative (bottom) results on noisy inputs with different level of visibility}
    \label{fig:occlude}
\end{figure}

As we can see from Figure~\ref{fig:occlude}, the errors (CD and EMD) increase only gradually as more and more regions are occluded. Note that our model is \emph{not} trained with these occluded and noisy examples, but it is still robust to them. The strong shape prior that the model has learned helps it to ignore the noisy points and predict reasonable outputs under occlusions. This in part explains its strong generalizability to real-world data, as we will show in the following section.

\subsection{Completion Results on KITTI}    \label{sec:kitti}
In this experiment, we test our completion method on partial point clouds from real-world LiDAR scans. Specifically, we take a sequence of Velodyne scans from the KITTI dataset \cite{geiger2013vision}. For each frame, we extract points within the object bounding boxes labeled as cars, which results in 2483 partial point clouds. Each point cloud is then transformed to the box's coordinates, completed with a PCN trained on cars from ShapeNet, and transformed back to the world frame. The process is illustrated in Figure~\ref{fig:kitti}. We use a model trained specifically on cars here to incorporate prior knowledge of the object class. Having such prior knowledge is not necessary for our method but will help the model achieve better performance.

We do not have the complete ground truth point clouds for KITTI. Thus, we propose three alternative metrics to evaluate the performance of our model: 1) Fidelity, which is the average distance from each point in the input to its nearest neighbour in the output. This measures how well the input is preserved;
2) Minimal Matching Distance (MMD), which is the Chamfer Distance (CD) between the output and the car point cloud from ShapeNet that is closest to the output point cloud in terms of CD. This measures how much the output resembles a typical car;
3) Consistency, which is the average CD between the completion outputs of the same instance in consecutive frames. This measures how consistent the network's outputs are against variations in the inputs. As a comparison, we also compute the average CD between the inputs in consecutive frames, denoted as Consistency (input).
These metrics are reported in Table~\ref{tab:kitti}. 

\begin{table}
    \centering
    \caption{Quantitative results on KITTI.}
    \begin{tabular}{c|c|c|c}
        \toprule
        Fidelity & MMD & Consistency & Consistency (input)  \\
        \midrule
        0.02800 & 0.01850 & 0.01163 & 0.05121 \\
        \bottomrule
    \end{tabular}
    \label{tab:kitti}
\end{table}

Unlike point clouds back-projected from 2.5D images, point clouds from LiDAR scans are very sparse. The 2483 partial point clouds here contain 440 points on average, with some having fewer than 10 points. In contrast, point clouds from 2.5D images used in training usually contain more than 1000 points. In spite of this, our model is able to transfer easily between the two distributions without any fine tuning, producing consistent completions from extremely partial inputs. This can be attributed to the use of point-based representation, which is less sensitive to input density than volumetric representations. In addition, each prediction with our model takes only 0.0012s on a Nvidia GeForce 1080Ti GPU and 2s on a 3.60GHz Intel Core i7-7700 CPU, making it suitable for real-time applications.

\begin{figure*}[ht]
  \begin{subfigure}[b]{0.33\textwidth}
    \includegraphics[width=\textwidth]{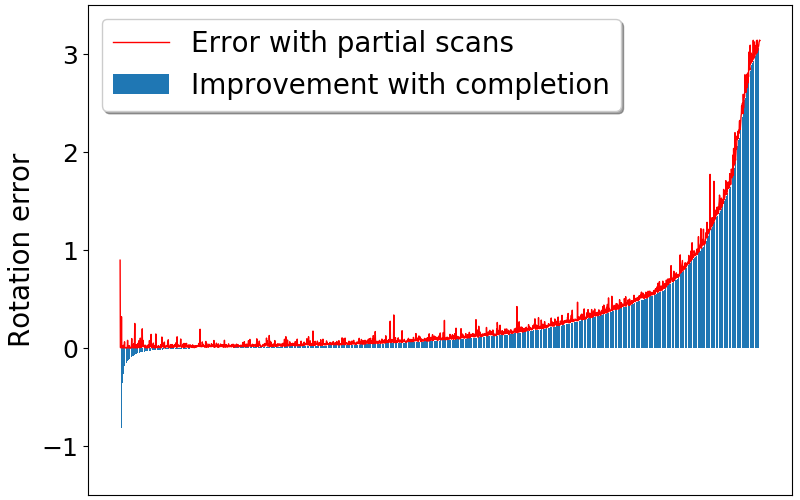}
    \caption{Rotation error}
    \label{fig:rot_err}
  \end{subfigure}
  \hfill
  \begin{subfigure}[b]{0.33\textwidth}
    \includegraphics[width=\textwidth]{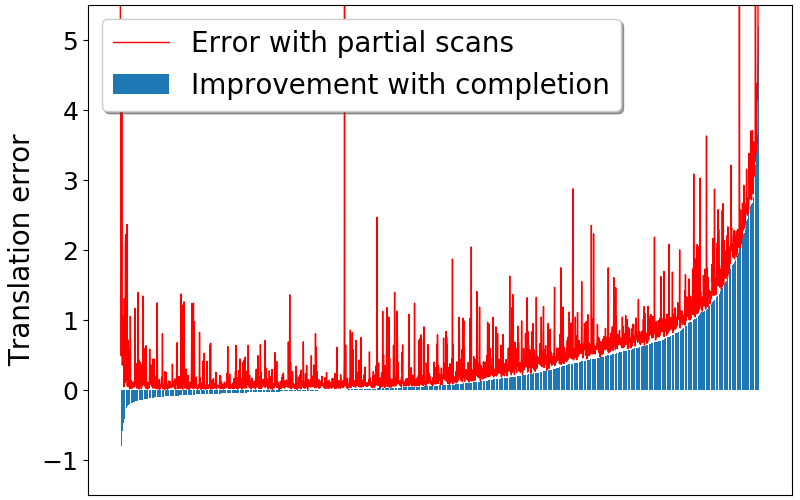}
    \caption{Translation error}
    \label{fig:trans_err}
  \end{subfigure}
  \hfill
  \begin{subfigure}[b]{0.33\textwidth}
    \includegraphics[width=\textwidth]{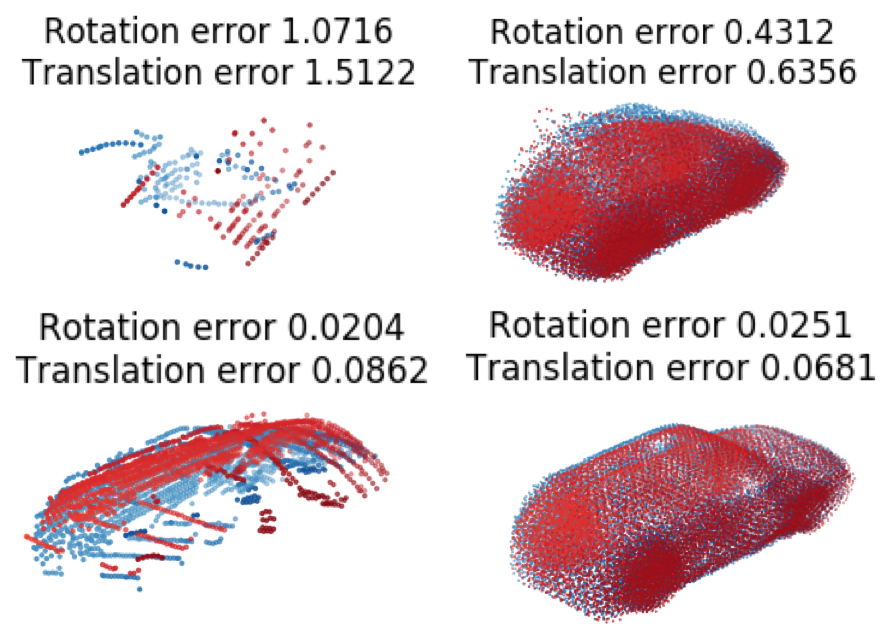}
    \caption{Qualitative example}
    \label{fig:regis}
  \end{subfigure}
  \caption{\textbf{Improvements on point cloud registration}. In (a) and (b), the $x$-axis represents different registration instances. The height of the blue bar indicates the amount of improvement of registration with complete point clouds over registration with partial point clouds. The red curve is the error of registration with partial point clouds and the difference between the red curve and the blue bar is the error of registration with complete point clouds. In (c), registered partial point clouds are shown on the left and registered complete point clouds of the same instances are shown on the right.}
\end{figure*}

\subsection{Point Cloud Registration with Completion}
Many common tasks on point clouds can benefit from a more complete and denser input. Here, as an example of such applications, we show that the output of our network can improve the results of point cloud registration. Specifically, we perform registration between car point clouds from neighboring frames in the same Velodyne sequence from Section \ref{sec:kitti}, using a simple point-to-point ICP \cite{besl1992method} algorithm implemented in PCL \cite{rusu20113d}.
This results in 2396 registration instances. We provide two kinds of inputs to the registration algorithm -- one is partial point clouds from the original scans, another is completed point clouds by a PCN trained on cars from ShapeNet. We compare the rotational and translational error on the registration results with partial and complete inputs. The rotational error is computed as $2\cos^{-1}(2\langle q_1, q_2\rangle^2-1)$, where $q_1$ and $q_2$ are the quaternion representations of the ground truth rotation and the rotation computed by ICP. This measures the angle between $q_1$ and $q_2$. The translational error is computed as $\norm{t_1-t_2}_2$, where $t_1$ is the ground truth translation and $t_2$ is the translation computed by ICP.

As shown in Figure~\ref{fig:rot_err} and \ref{fig:trans_err}, both rotation and translation estimations are more accurate with complete point clouds produced by PCN, and the improvement is most significant when the error with partial point clouds is large. Figure~\ref{fig:regis} shows some qualitative examples. As can be seen, the complete point clouds are much easier to register because they contain larger overlaps, a lot of which are regions completed by PCN.
Note that the improvement brought by our completion results is not specific to ICP, but can be applied to any registration algorithm. 

\section{Discussion}
\begin{figure}[ht]
    \centering
    \includegraphics[width=\linewidth]{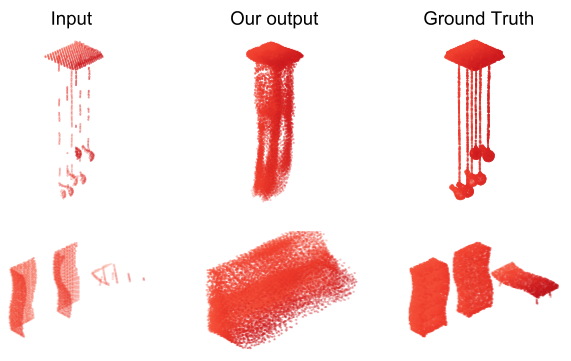}
    \caption{\textbf{Failure modes}: thin structures (top) and disconnected parts (bottom).}
    \label{fig:failures}
\end{figure}

We have identified two prominent failure modes for our model. 
First, there are some object instances consisting of multiple disconnected parts. 
Our model fails to recognize this and incorrectly connects the parts. This is likely a result of the strong priors learned from the training dataset where almost all objects are connected.
Second, some objects contain very thin structures such as wires. Our model is occasionally unable to recover these structures. There are two possible reasons. First, the points from these structures are often sparse since they have small surface areas, which makes the 3D feature extraction more difficult. Second, unlike most object surfaces the local geometry of thin structures does not resemble the 2D grid, making it challenging for our model to deform a 2D grid into these thin structures. Some visualizations of these failures are shown in Figure \ref{fig:failures}.

\begin{figure*}[ht]
    \centering
    \includegraphics[width=\linewidth]{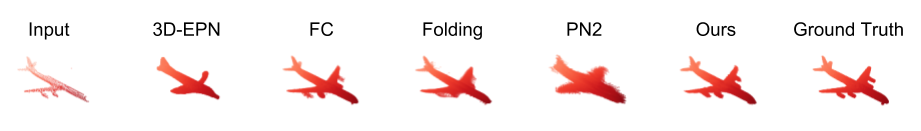}
    \includegraphics[width=\linewidth]{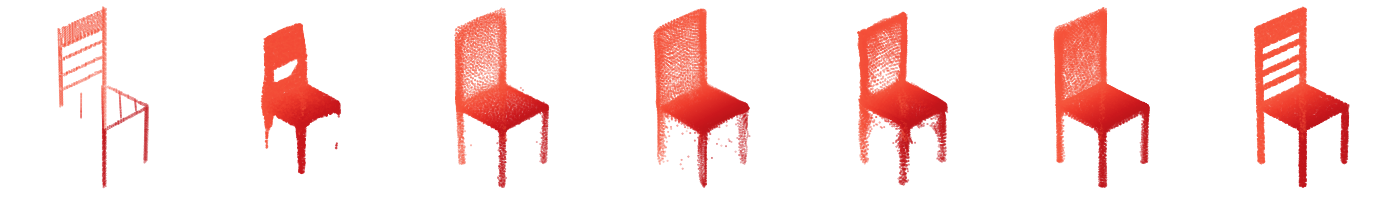}
    \includegraphics[width=\linewidth]{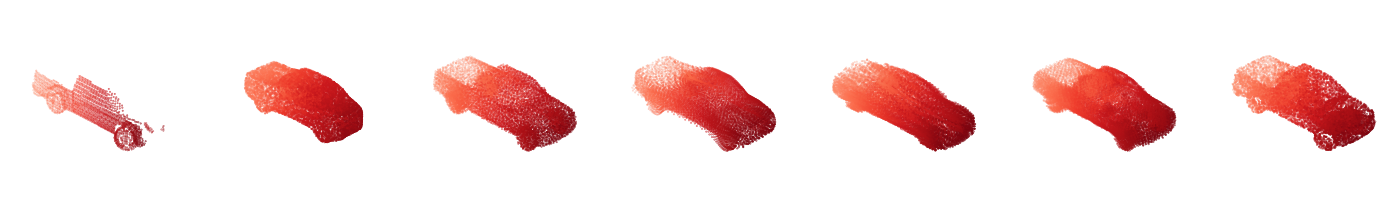}
    \includegraphics[width=\linewidth]{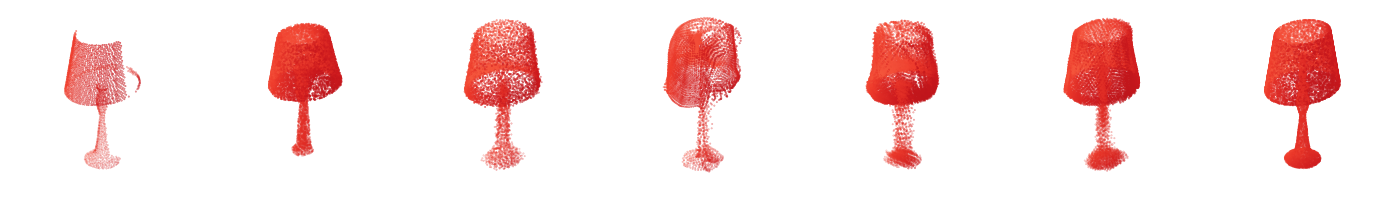}
    \noindent\rule{\linewidth}{0.4pt}
    \includegraphics[width=\linewidth]{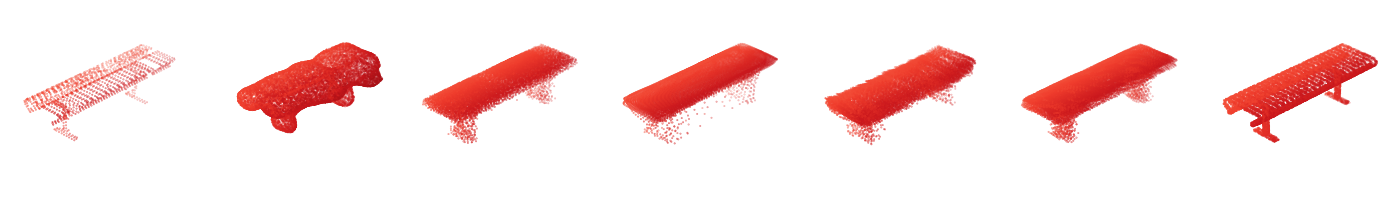}
    \includegraphics[width=\linewidth]{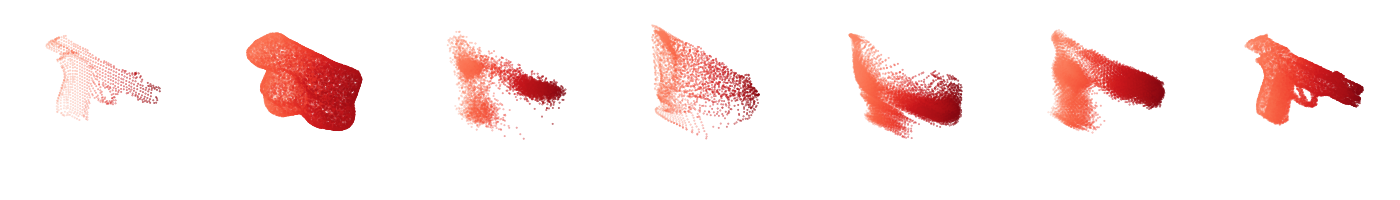}
    \includegraphics[width=\linewidth]{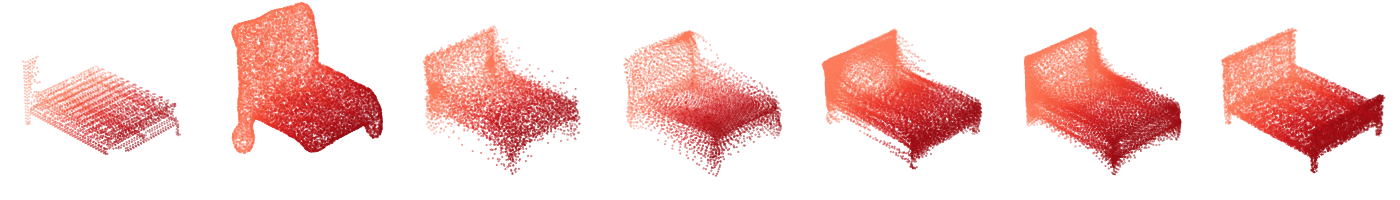}
    \includegraphics[width=\linewidth]{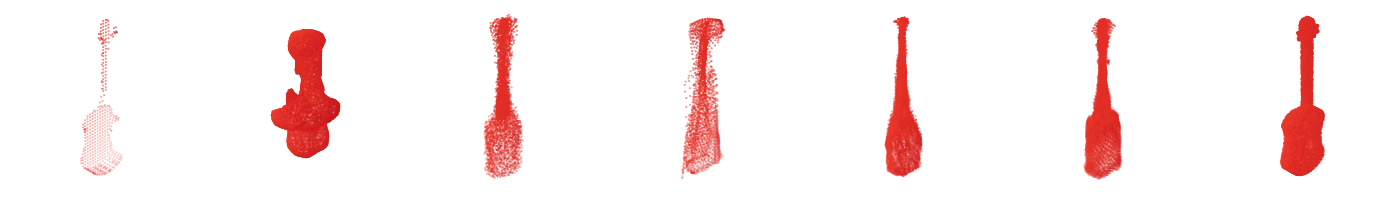}
    \caption{\textbf{Qualitative completion results on ShapeNet}. Top four rows are results on categories used during training. Bottom four rows are results on categories not seen during training.}
    \label{fig:qualitative}
\end{figure*}

\section{Conclusion}
We have presented a new approach to shape completion using point clouds without any voxelization. To this end, we have designed a deep learning architecture which combines advantages from existing architectures to generate a dense point cloud in a coarse-to-fine fashion, enabling high resolution completion with much fewer parameters than voxel-based models. Our method is effective across multiple object categories and works with inputs from different sensors. In addition, it shows strong generalization performance on unseen objects and real-world data. Our point-based completion method is more scalable and robust than voxel-based methods, which makes it a better candidate for completion of more complex data such as scenes.

\section*{Acknowledgements}
This project is supported by Carnegie Mellon University's Mobility21 National University Transportation Center, which is sponsored by the US Department of Transportation. We would also like to thank Adam Harley, Leonid Keselman and Rui Zhu for their helpful comments and suggestions.

{\small
\bibliographystyle{ieee}
\bibliography{main}}
\clearpage

\renewcommand\thesection{\Alph{section}}
\renewcommand\thesubsection{\thesection.\Alph{subsection}}
\setcounter{section}{0}
\section{Overview}
In this document we provide technical details and additional quantitative and qualitative results to the main paper.

In Section \ref{supp:folding}, we describe the local folding operation in detail. Section \ref{supp:network} provides specific parameters for the models compared in Section \ref{sec:shapenet}. Section \ref{supp:shapenet} and \ref{supp:kitti} present more results on ShapeNet and KITTI, including failure cases. Section \ref{supp:analysis} provides further analysis on the network design. Section \ref{supp:visu} shows more visualization results.

\section{Local Folding Operation} \label{supp:folding}
\begin{figure}[ht]
    \centering
    \includegraphics[width=\linewidth]{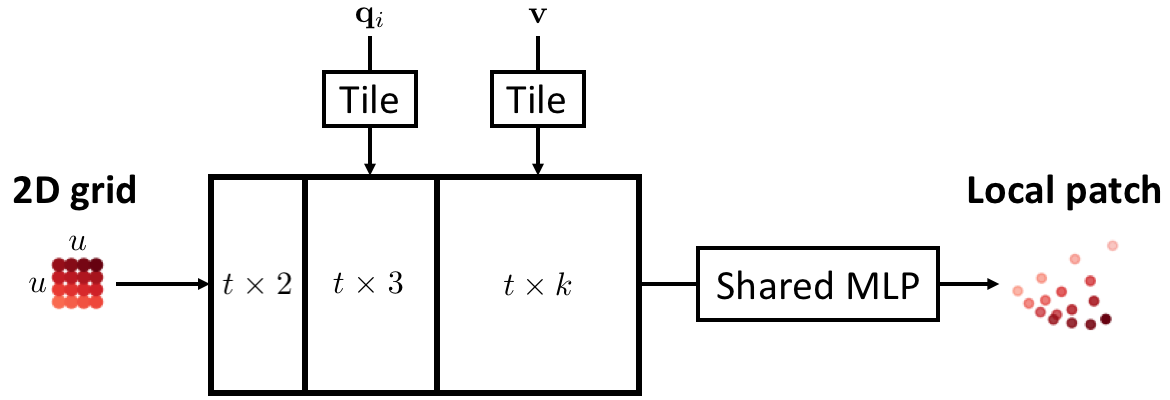}
    \caption{The folding operation}
    \label{fig:folding}
\end{figure}

Here we describe the local folding operation mentioned in Section \ref{sec:decoder} in detail. As shown in Figure \ref{fig:folding}, the folding operation takes a point $\mathbf{q}_i$ in the coarse output $Y_{coarse}$ and the $k$-dimensional global feature $\mathbf{v}$ as inputs, and generates a patch of $t=u^2$ points in local coordinates centered at $\mathbf{q}_i$ by deforming a $u\times u$ grid. It first takes points on a zero-centered $u\times u$ grid with side length $r$ ($r$ controls the scale of the output patch) and organize their coordinates into a $t\times2$ matrix $G$. Then, it concatenates each row of $G$ with the coordinates of the center point $\mathbf{q}_i$ and the global feature vector $\mathbf{v}$, and passes the resulting matrix through a shared MLP that generates a $t\times 3$ matrix $Q$, i.e. the local patch centered at $\mathbf{q}_i$. This shared MLP can be interpreted as a non-linear transformation that deforms the 2D grid into a smooth 2D manifold in 3D space. Note that the same MLP is used in the local patch generation for each $\mathbf{q}_i$ so the number of parameters in the local folding operation does not grow with the output size.

\section{Network Architecture Details} \label{supp:network}
Here we describe the detailed architecture of the models compared in Section \ref{sec:shapenet}. The 3D-EPN model we used is 3D-EPN-unet-class, the best performing model from \cite{dai2017shape}. For the seen categories, we use the $128^3$ output which involves an additional database retrieval step. For the unseen categories, we use the original $32^3$ output from the model.

The stacked PN encoder used by FC, Folding, PCN-CD and PCN-EMD includes 2 PN layers. The shared MLP in the first PN layer has 2 layers with 128 and 256 units. The shared MLP in the second PN layer has 2 layers with 512 and 1024 units. The PN2 encoder follows the same architecture as the SSG network in \cite{qi2017pointnet++}.

The FC decoder contains 3 fully-connected layers with 1024, 1024 and $16384\cdot3$ units. The Folding decoder contains 2 folding layers as in \cite{yang2017foldingnet}, where the second layer takes the output of the first layer instead of a 2D grid. Note that these folding layers are different from the one described in Section \ref{supp:folding} in that they do not take the center point coordinates as input. Each folding layer has a 3-layer shared MLP with 512, 512 and 3 units. The grid size is $u=128$ and the grid scale is $r=0.5$.

The multistage decoder in PCN-CD and PCN-EMD has 3 fully-connected layers with 1024, 1024 and $1024\cdot3$ units, followed by 1 folding layer as described in Section \ref{supp:folding}, where the grid size is $u=4$ and the grid scale is $r=0.05$. The folding layer contains a 3-layer shared MLP with 512, 512 and 3 units.

\begin{figure*}[ht]
    \centering
    \includegraphics[width=0.92\columnwidth]{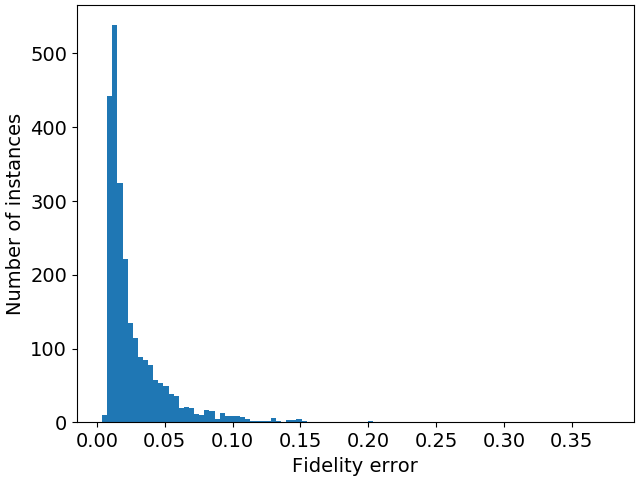}
    \includegraphics[width=0.92\columnwidth]{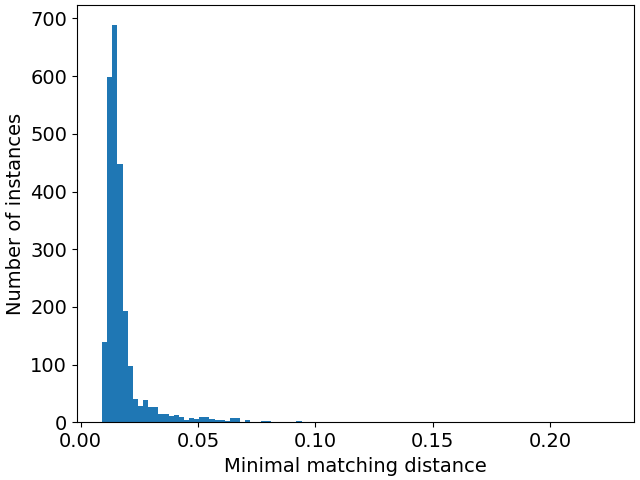}
    \caption{Distribution of fidelity error and minimal matching distance on KITTI car completions}
    \label{fig:hist}
\end{figure*}

\section{Additional Results on ShapeNet} \label{supp:shapenet}
Table \ref{tab:cd_seen}, \ref{tab:emd_seen}, \ref{tab:cd_unseen} and \ref{tab:emd_unseen} show the quantitative results on test instances from ShapeNet corresponding to Figure \ref{fig:comparison} in the main paper. Figure \ref{fig:qualitative} shows the qualitative comparisons on shapes from seen as well as unseen categories. As can be seen, the outputs of 3D-EPN often contain missing or extra parts. The outputs of FC are accurate but the points are overly concentrated in certain regions. The outputs of Folding contain many floating points and the outputs of PN2 are blurry. The outputs of our model best match the ground truth in terms of global geometry and local point density.


\section{Additional Results on KITTI} \label{supp:kitti}
\begin{figure}[ht]
    \centering
    \includegraphics[width=\linewidth]{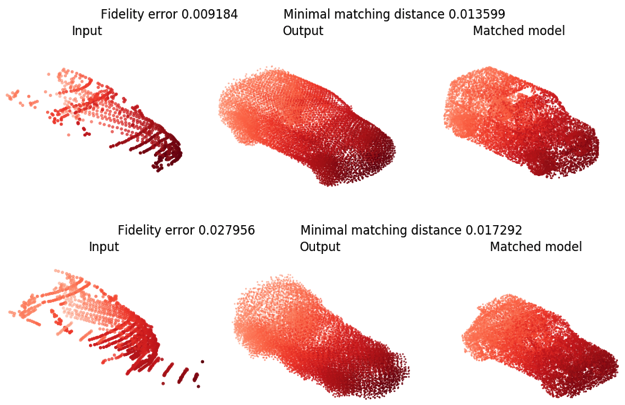}
    \caption{\textbf{Completion (middle) and matched model (right) for cars in KITTI (left)}. The matched model on the right is the car point cloud from ShapeNet that is closest to the completion output in Chamfer Distance. Top figure shows a successful completion and bottom figure shows a failure case caused by incorrect segmentation.}
    \label{fig:match}
\end{figure}

Figure \ref{fig:hist} shows the distribution of fidelity error and minimal matching distance on the completion results on KITTI. It can be seen that there are a few failure cases with very high error that can bias the mean value reported in Section \ref{sec:kitti}. Figure \ref{fig:match} shows some qualitative examples. We observe that in most cases, our model produces a valid car shape that matches the input while being different from the matched model in ShapeNet, as the one shown in the top figure. However, we also observe some failure cases, e.g. the one shown in the bottom figure, caused by extra points from the ground or nearby objects that are within the car's bounding box. This problem can potentially be resolved by adding a segmentation step before passing the partial point cloud to PCN.

\section{More Architecture Analysis} \label{supp:analysis}
\begin{figure}[ht]
    \centering
    \includegraphics[width=\linewidth]{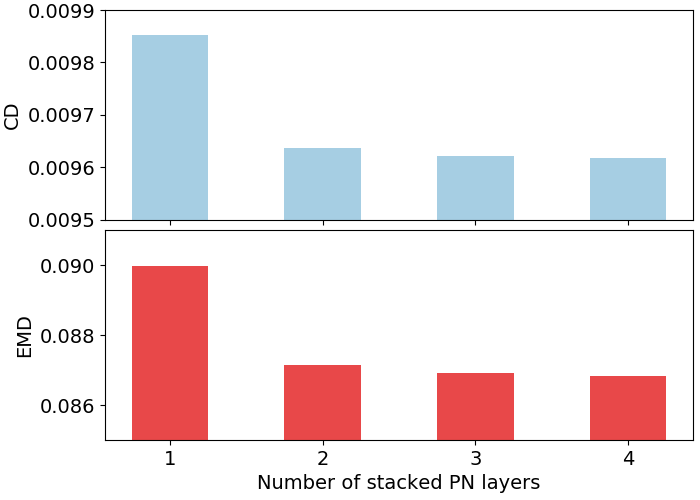}
    \caption{Test error with different number of PN layers}
    \label{fig:stacked}
\end{figure}

\paragraph{Effect of stacked PN layers} Here we test variants of PCN-CD with different number of stacked PN layers in the encoder. The mean CD and EMD on the ShapeNet test set are shown in Figure \ref{fig:stacked}. It can be seen that the advantage of using 2 stacked PN layers over 1 is quite apparent, whereas the benefit of using more stacked PN layers is almost negligible. This shows that 2 stacked PN layers are sufficient for mixing the local and global geometry information in the input. Thus, we keep the number of stacked PN layers as 2 in our experiments, even though PCN's performance can be further improved by using more stacked PN layers.

\paragraph{Effect of bottleneck size}
\begin{figure}[ht]
    \centering
    \includegraphics[width=\linewidth]{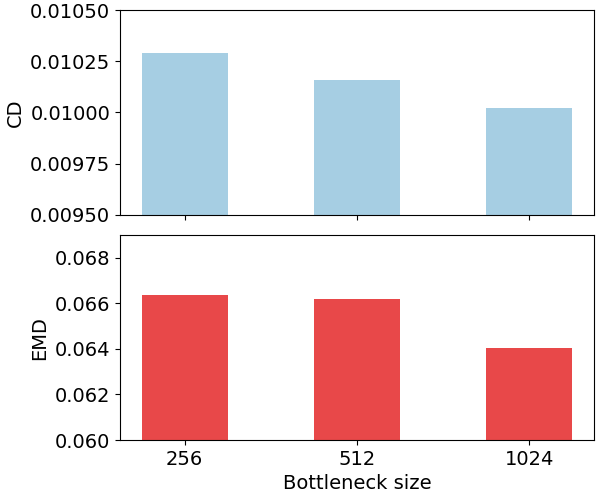}
    \caption{Test error with different bottleneck sizes}
    \label{fig:bottleneck}
\end{figure}
Here we test variants of PCN-EMD with different bottleneck size, i.e. the length of the global feature vector $\mathbf{v}$. The mean CD and EMD on the ShapeNet test set are shown in Figure \ref{fig:bottleneck}. It can be seen that PCN's performance improves as the bottleneck size increases. In our experiments, we choose the bottleneck size to be 1024 because a larger bottleneck of size 2048 cannot fit into the memory of a single GPU. This implies that if multiple GPUs are used for training, PCN's performance can further improve with a larger bottleneck.

\section{More Visualizations} \label{supp:visu}
\paragraph{Keypoint Visualization}
As noted in \cite{qi2017pointnet}, the PN layer can be interpreted as selecting a set of keypoints which describes the shape. More specifically, each output unit of the shared MLP can be considered as a function on the points. The keypoints are points that achieve the maximum value for at least one of these point functions. In other words, they are points whose feature values are ``selected" by the maxpooling operation to be in the final feature vector.
Note that a point can be selected more than once by achieving the maximum for multiple feature functions. In fact, as shown in Table \ref{tab:keypoint}, the number of keypoints is usually far less than the bottleneck size (1024) or the number of input points. This implies that as long as these keypoints are preserved, the learned features won't change. This property contributes to our model's robustness against noise.

\begin{figure}[ht]
    \centering
    \includegraphics[width=1.05\linewidth]{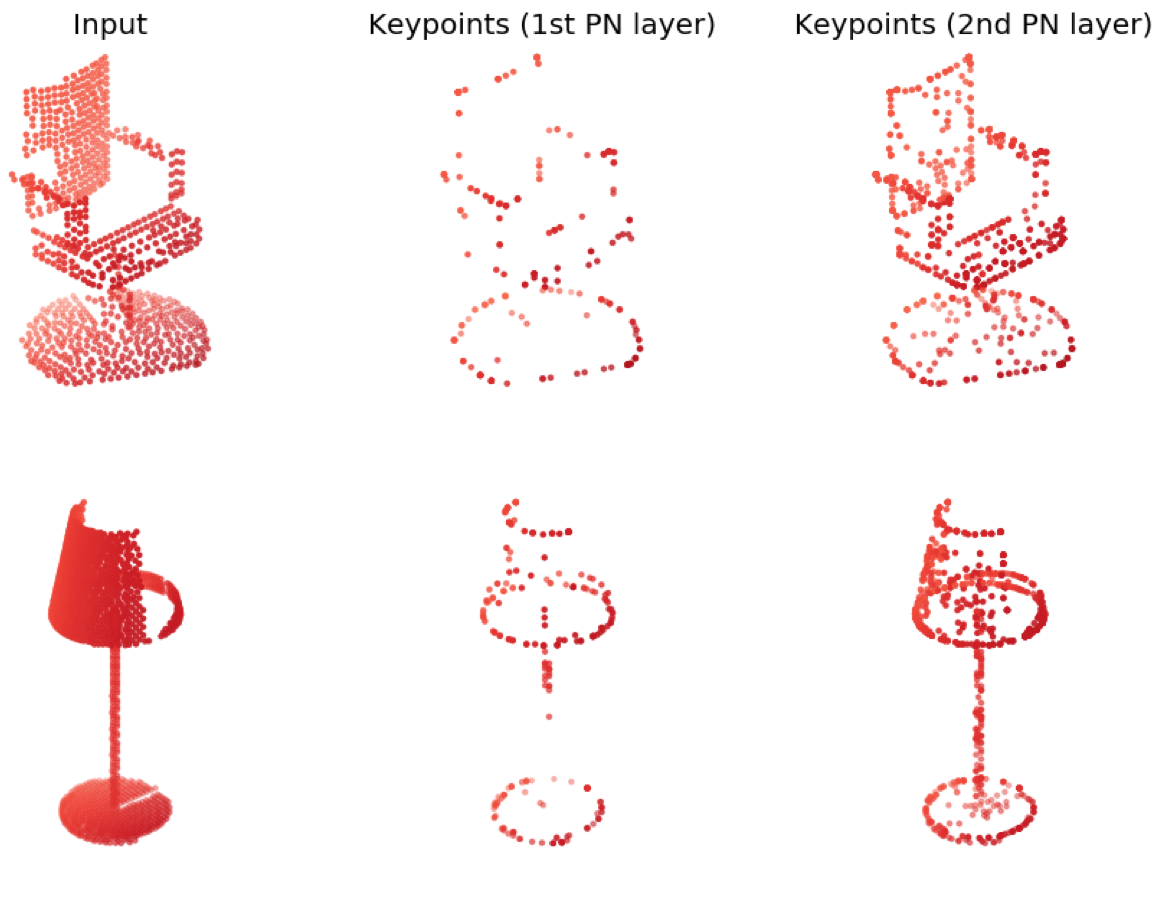}
    \caption{Keypoints visualization}
    \label{fig:keypoints}
\end{figure}

In Figure \ref{fig:keypoints}, we visualize the keypoints selected by the two PN layers in our stacked PN encoder. It can be observed that the two PN layers summarizes the shape in a coarse-to-fine fashion -- the first layer selects just a few points that compose the outline of the shape, while the second layer select more points that further delineate the visible surfaces. Note that this coarse-to-fine description emerges without any explicit supervision.

\begin{table}[ht]
    \centering
    \begin{tabular}{c|c}
        \toprule
         & Average number of points \\ \midrule
        Input & 1105 \\
        Keypoints (1st PN layer) & 101 \\
        Keypoints (2nd PN layer) & 363 \\
        \bottomrule
    \end{tabular}
    \caption{Number of keypoints versus number of input points}
    \label{tab:keypoint}
\end{table}

\begin{figure*}[ht]
    \centering
    \includegraphics[width=\linewidth,height=\textheight, keepaspectratio]{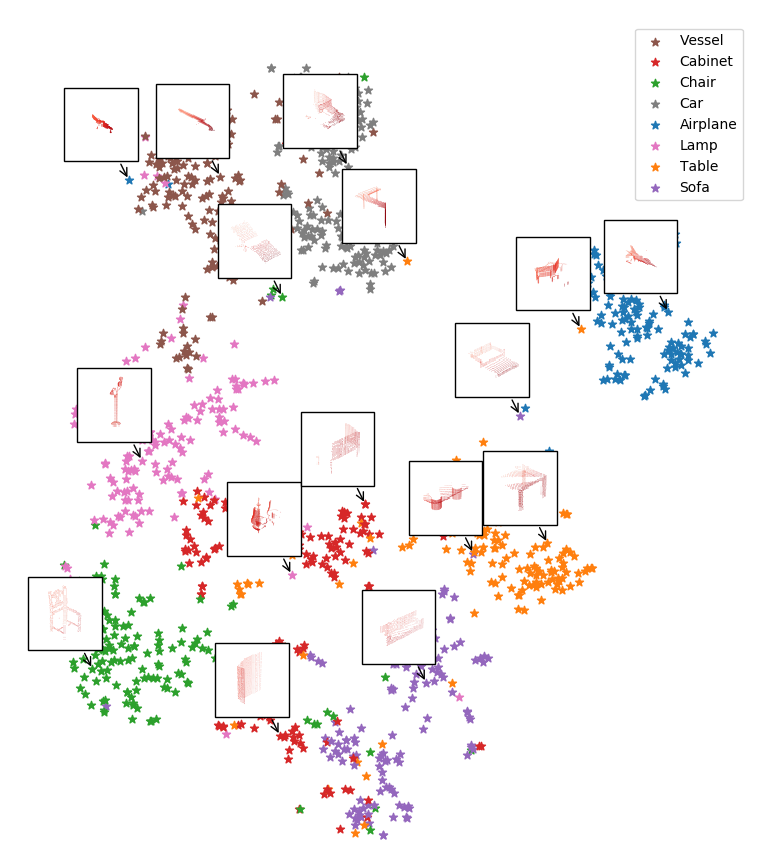}
    \caption{T-SNE embedding of learned features on partial point clouds}
    \label{fig:tsne}
\end{figure*}

\paragraph{Feature Space Visualization}
In Figure \ref{fig:tsne}, we use t-SNE \cite{maaten2008visualizing} to embed the 1024-dimensional global features of the ShapeNet test instances into a 2D space. It can be seen that shapes are clustered together by their semantic categories. Note that this is also an emerging behavior without any supervision, since we do not use the category labels at all during training.

\begin{figure*}
    \centering
    \includegraphics[width=\columnwidth]{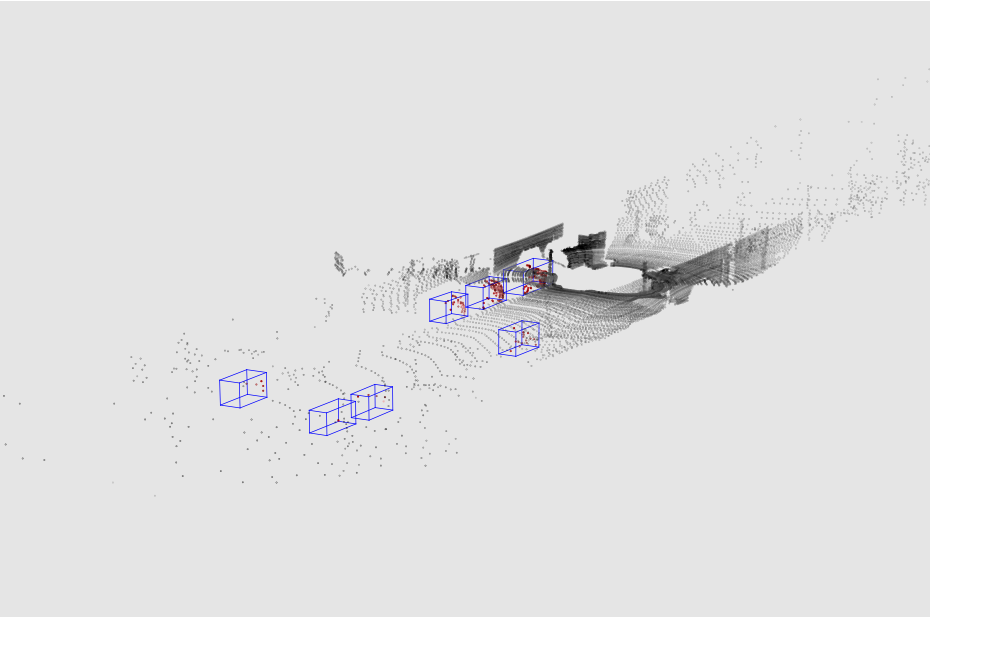}
    \includegraphics[width=\columnwidth]{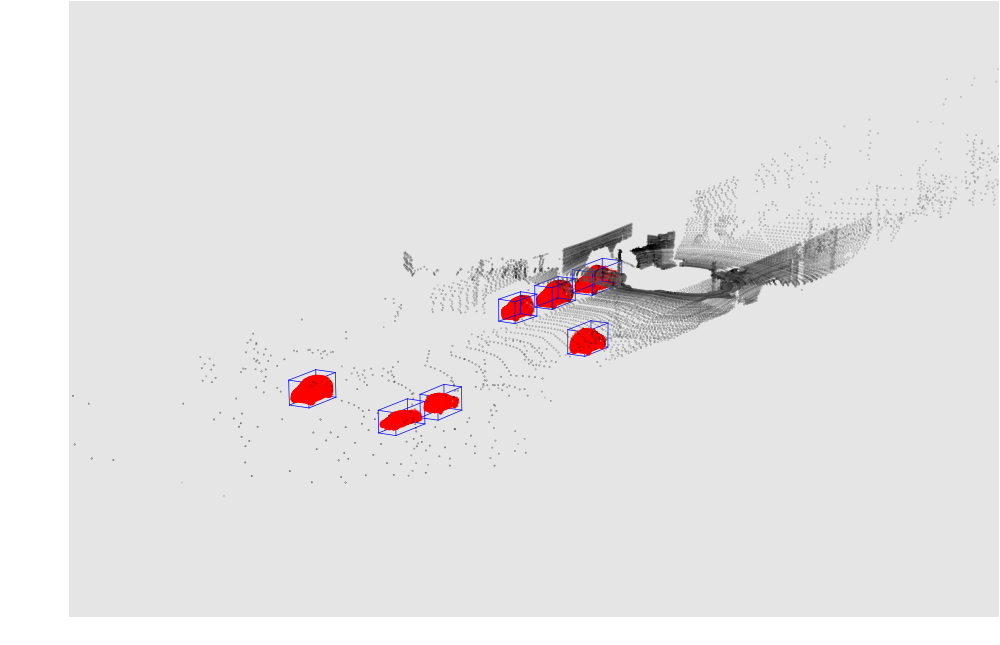}
    \includegraphics[width=\columnwidth]{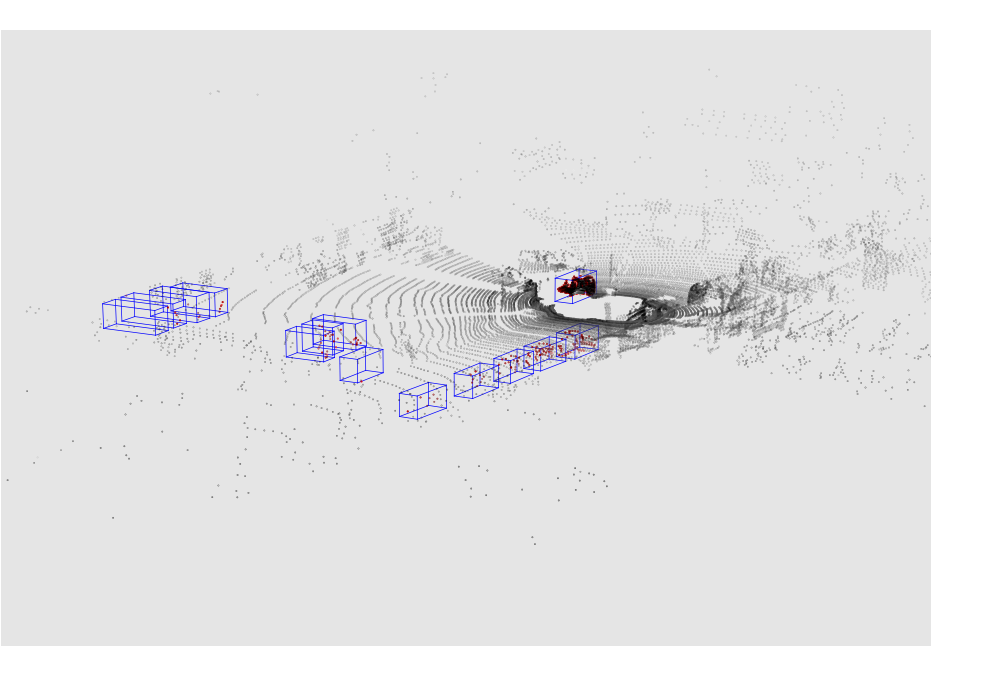}
    \includegraphics[width=\columnwidth]{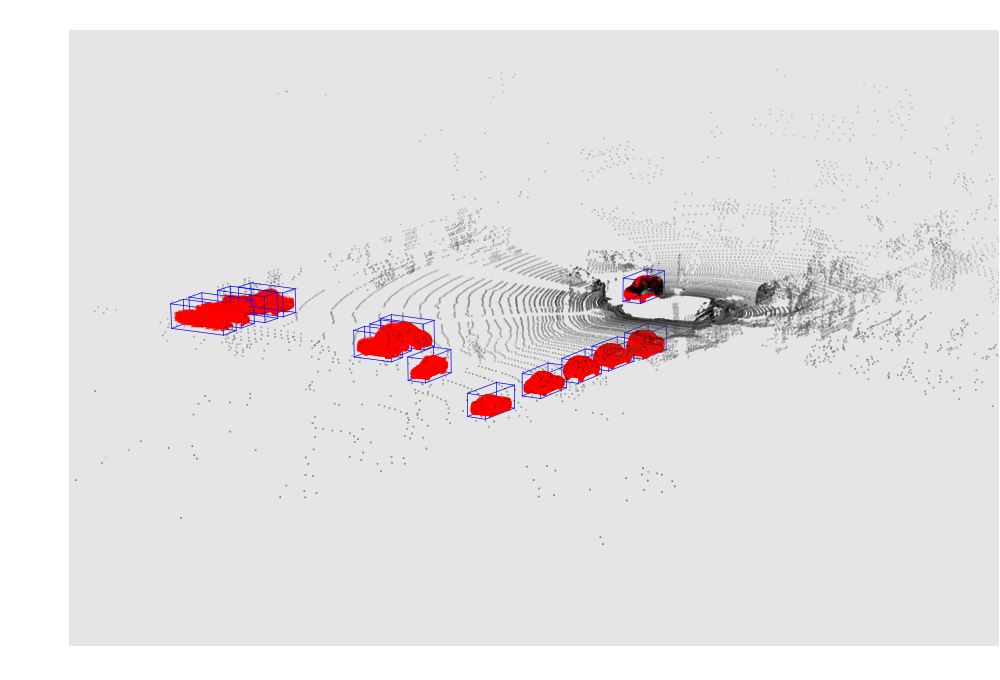}
    \includegraphics[width=\columnwidth]{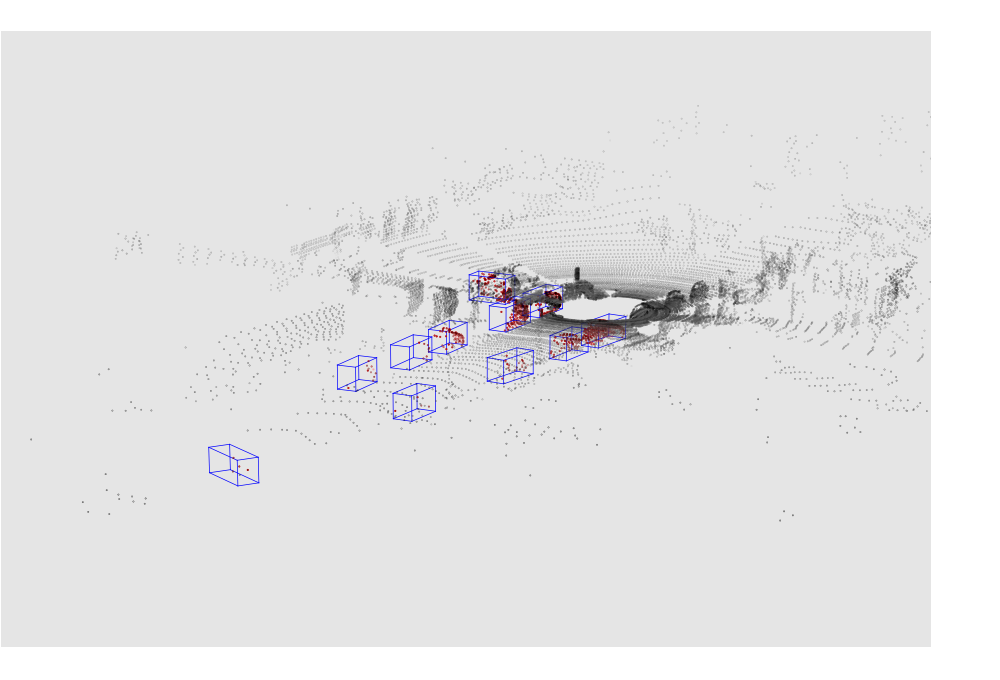}
    \includegraphics[width=\columnwidth]{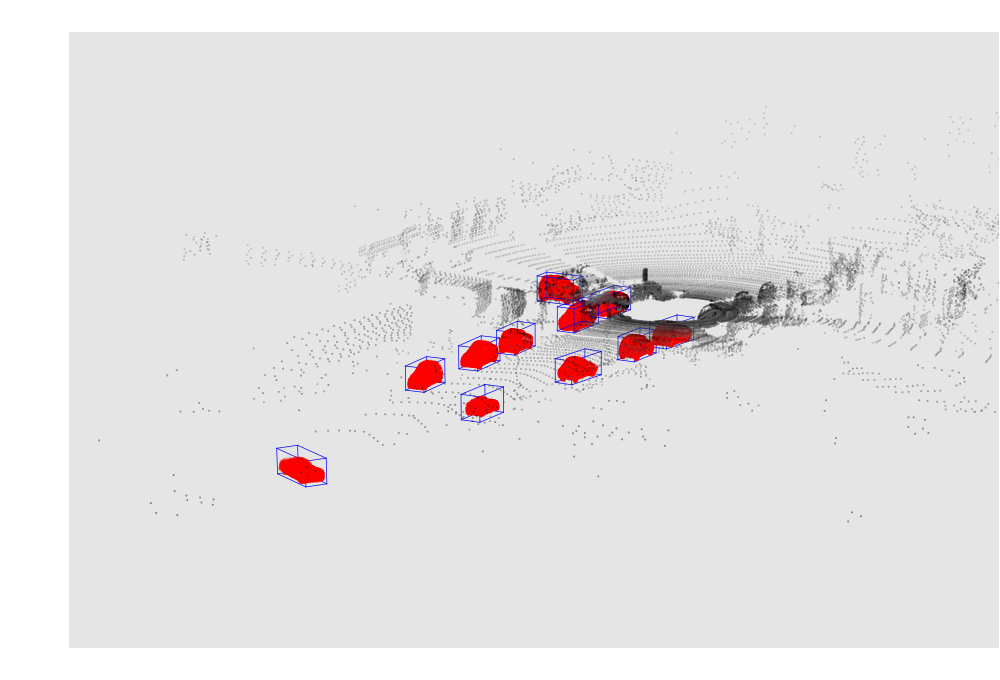}
    \includegraphics[width=\columnwidth]{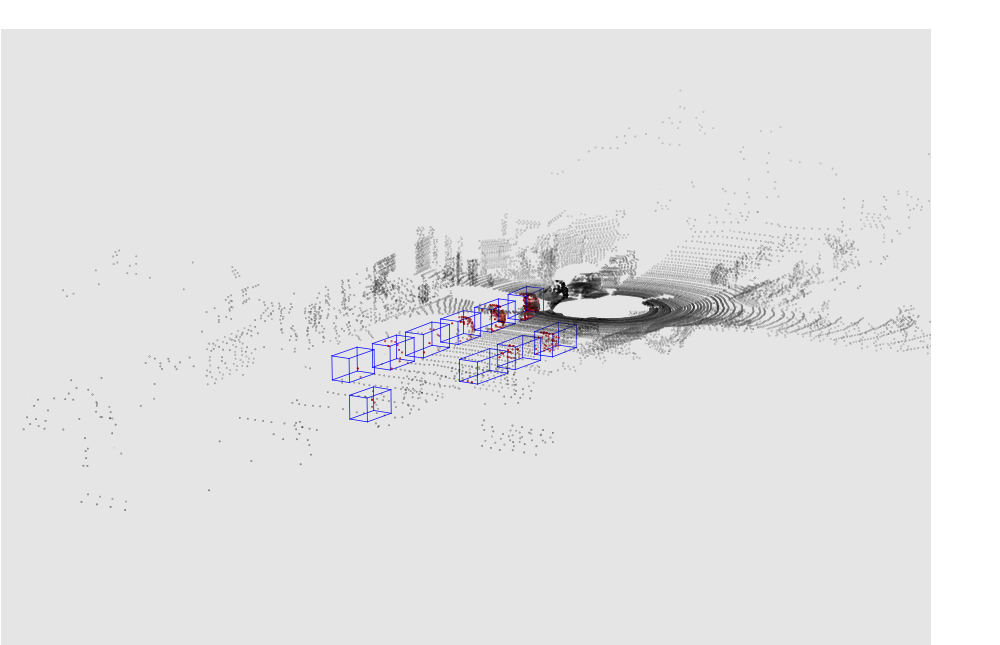}
    \includegraphics[width=\columnwidth]{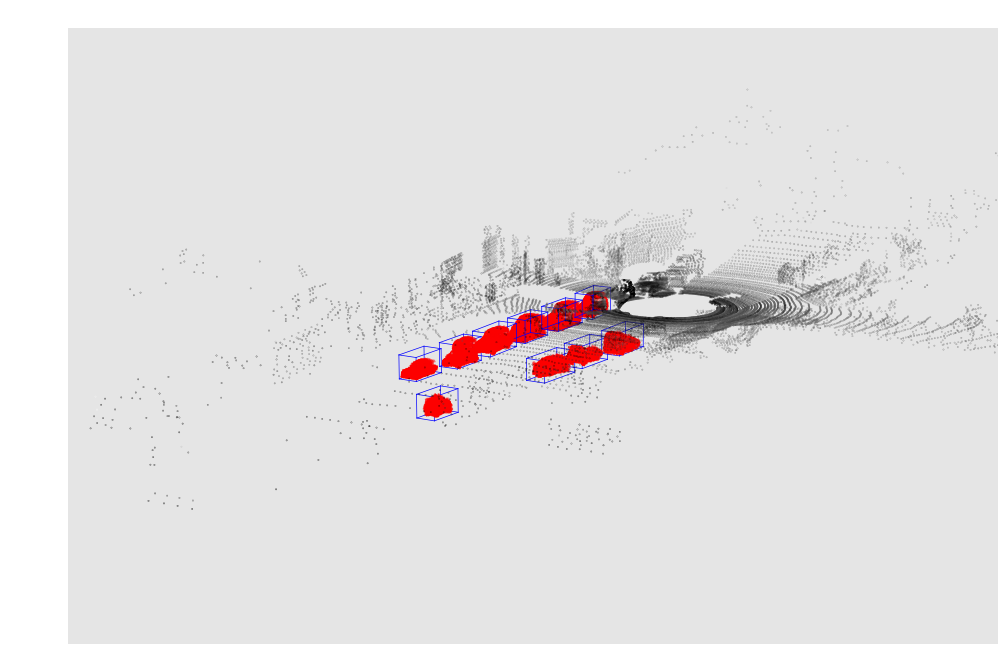}
    \caption{Qualitative completion results on KITTI}
    \label{fig:kitti_scan}
\end{figure*}

\begin{table*}[h]
\centering
\caption{Seen categories of ShapeNet dataset - Chamfer Distance}
\label{tab:cd_seen}
\begin{tabular}{@{}l|l|llllllll@{}}
\hline
\multirow{2}{*}{Method}          & \multicolumn{9}{c}{Mean Chamfer Distance per point}                         \\ \cmidrule(l){2-10}
                          & Avg               & Airplane & Cabinet & Car & Chair & Lamp & Sofa & Table & Vessel \\ \cmidrule(r){1-10}
3D-EPN  & 0.020147          & 0.013161 & 0.021803 & 0.020306 & 0.018813 & 0.025746 & 0.021089 & 0.021716 & 0.018543 \\ 
FC                    & 0.009799          & 0.005698 & 0.011023 & 0.008775 & \textbf{0.010969} & \textbf{0.011131} & 0.011756 & 0.009320 & 0.009720 \\ 
Folding               & 0.010074          & 0.005965 & 0.010831 & 0.009272 & 0.011245 & 0.012172 & 0.011630 & 0.009453 & 0.010027 \\ 
PN2             & 0.013999          & 0.010300 & 0.014735 & 0.012187 & 0.015775 & 0.017615 & 0.016183 & 0.011676 & 0.013521 \\
PCN-CD              & \textbf{0.009636} & \textbf{0.005502} & \textbf{0.010625} & \textbf{0.008696} & 0.010998 & 0.011339 & \textbf{0.011676} & \textbf{0.008590} & \textbf{0.009665} \\ 
PCN-EMD             & 0.010021          & 0.005849 & 0.010685 & 0.009080 & 0.011580 & 0.011961 & 0.012206 & 0.009014 & 0.009789 \\ \bottomrule
\end{tabular}
\end{table*}

\begin{table*}[h]
\centering
\caption{Seen categories of ShapeNet dataset - Earth Mover's Distance}
\label{tab:emd_seen}
\begin{tabular}{@{}l|l|llllllll@{}}
\hline
\multirow{2}{*}{Method}          & \multicolumn{9}{c}{Mean Earth Mover's Distance per point}                         \\ \cmidrule(l){2-10}
                          & Avg               & Airplane & Cabinet & Car & Chair & Lamp & Sofa & Table & Vessel \\ \cmidrule(r){1-10}
3D-EPN  & 0.081785   & 0.061960 & 0.077630 & 0.087044 & 0.076802 & 0.107317 & 0.080802 & 0.080996 & 0.081732 \\ 
FC      & 0.171280          & 0.073556 & 0.214723 & 0.157297 & 0.189727 & 0.240547 & 0.191488 & 0.161117 & 0.141782 \\ 
Folding & 0.228015          & 0.156438 & 0.221349 & 0.174567 & 0.297427 & 0.319983 & 0.245664 & 0.189904 & 0.218788 \\ 
PN2     & 0.101445          & 0.059574 & 0.116179 & 0.066942 & 0.110595 & 0.185817 & 0.102642 & 0.086053 & 0.083755 \\
PCN-CD  & 0.087142          & 0.046637 & 0.097691 & 0.057178 & 0.086787 & 0.169540 & 0.083425 & 0.080783 & 0.075094 \\ 
PCN-EMD & \textbf{0.064044} & \textbf{0.038752} & \textbf{0.070729} & \textbf{0.054967} & \textbf{0.068074} & \textbf{0.084613} & \textbf{0.072437} & \textbf{0.060069} & \textbf{0.062713} \\ \bottomrule
\end{tabular}
\end{table*}

\begin{table*}[h]
\centering
\caption{Unseen categories of ShapeNet dataset - Chamfer Distance}
\label{tab:cd_unseen}
\begin{tabular}{@{}l|lllll|lllll@{}}
\hline
\multirow{2}{*}{Method}          & \multicolumn{10}{c}{Mean Chamfer Distance per point}                         \\ \cmidrule(l){2-11}
                          & Avg & Bus & Bed & Bookshelf & Bench & Avg & Guitar & Motorbike & Skateboard & Pistol \\ \cmidrule(r){1-11}
3D-EPN      & 0.0415 & 0.03594 & 0.04785 & 0.03912 & 0.04307 & 0.0443 & 0.04735 & 0.04067 & 0.04784 & 0.04136\\
FC          & 0.0142 & 0.00982 & 0.02123 & 0.01512 & 0.01081 & 0.0129 & 0.00992 & \textbf{0.01456} & 0.01200 & 0.01497 \\
Folding     & \textbf{0.0138} & 0.01058 & \textbf{0.01908} & 0.01488 & \textbf{0.01055} & \textbf{0.0124} & \textbf{0.00906} & 0.01556 & \textbf{0.01191} & \textbf{0.01313} \\
PN2         & 0.0169 & 0.01260 & 0.02378 & 0.01687 & 0.01445 & 0.0168 & 0.01429 & 0.01635 & 0.01290 & 0.02353 \\
PCN-CD      & 0.0142 & \textbf{0.00946} & 0.02163 & \textbf{0.01479} & 0.01102 & 0.0129 & 0.01040 & 0.01475 & 0.01204 & 0.01423 \\
PCN-EMD     & 0.0146 & 0.00972 & 0.02236 & 0.01496 & 0.01139 & 0.0131 & 0.01147 & 0.01525 & 0.01211 & 0.01359 \\
\bottomrule
\end{tabular}
\end{table*}

\begin{table*}[h]
\centering
\caption{Unseen categories of ShapeNet dataset - Earth Mover's Distance}
\label{tab:emd_unseen}
\begin{tabular}{@{}l|lllll|lllll@{}}
\hline
\multirow{2}{*}{Method}          & \multicolumn{10}{c}{Mean Earth Mover's Distance per point}                         \\ \cmidrule(l){2-11}
                          & Avg & Bus & Bed & Bookshelf & Bench & Avg & Guitar & Motorbike & Skateboard & Pistol \\ \cmidrule(r){1-11}
3D-EPN      & 0.1189 & 0.10681 & 0.13318 & 0.11856 & 0.11711 & 0.1309 & 0.16255 & 0.11893 & 0.11328 & 0.12873 \\
FC          & 0.1998 & 0.16686 & 0.25567 & 0.20619 & 0.17050 & 0.1790 & 0.17635 & 0.18132 & 0.16706 & 0.19134 \\
Folding     & 0.2494 & 0.22150 & 0.32603 & 0.22555 & 0.22457 & 0.2733 & 0.32040 & 0.25137 & 0.25435 & 0.26689 \\
PN2         & 0.1062 & 0.08081 & 0.14900 & 0.12155 & 0.07349 & 0.1270 & 0.17836 & 0.10372 & 0.08825 & 0.13749 \\
PCN-CD      & 0.0908 & 0.06270 & 0.13556 & 0.10332 & 0.06161 & 0.1130 & 0.16834 & \textbf{0.09206} & 0.08464 & 0.10702 \\
PCN-EMD     & \textbf{0.0705} & \textbf{0.05991} & \textbf{0.10350} & \textbf{0.07607} & \textbf{0.06044} & \textbf{0.0816} & \textbf{0.07478} & 0.09471 & \textbf{0.06249} & \textbf{0.09426} \\
\bottomrule
\end{tabular}
\end{table*}

\end{document}